%% file: main.tex
\documentclass[11pt]{article}
\usepackage[margin=1in]{geometry}
\usepackage[T1]{fontenc}
\usepackage{lmodern}
\usepackage{amsmath,amssymb}
\usepackage{booktabs}
\usepackage{array}
\usepackage{graphicx}
\usepackage{algorithm}
\usepackage{algpseudocode}
\usepackage[numbers,sort&compress]{natbib}
\usepackage{xcolor}
\usepackage[hidelinks]{hyperref}
\graphicspath{{./}}

\title{Folding Attention Biases into Keys Makes Compacted KV Caches Run on Standard Kernels}
\author{Utkarsh Kanwat\\Independent Researcher\\\texttt{utkarshkanwat@gmail.com}}
\date{}

\begin{document}
\maketitle

\input{sections/abstract}
\input{sections/intro}
\input{sections/background}
\input{sections/method}
\input{sections/why}
\input{sections/downstream}
\input{sections/deploy}
\input{sections/ablations}
\input{sections/related}
\input{sections/limits}

\input{sections/conclusion}

\bibliographystyle{plainnat}
\renewcommand{\bibfont}{\small}
\setlength{\bibsep}{2pt plus 1pt}
\bibliography{refs}

\appendix
\input{sections/appendix}

\end{document}

%% file: sections/abstract.tex
\begin{abstract}
Attention Matching (AM) compacts a KV cache by keeping some keys per head, refitting their values and adding a bias to each kept key's logit.
FlashAttention and vLLM's paged caches have no input for it.
We fold the bias into the keys along one closed-form direction per head, fitted on context queries, giving a plain cache with AM's size and head dimension and the model's own query path; the output error is bounded by how far each query's projection on that direction is from 1.
With equal per-head budgets, unmodified vLLM serves folded caches through a KV connector, matching Hugging Face's answers on all 51 Qwen3-4B questions and its greedy output on 180 of 200 Qwen3-14B LongBench samples.
With equal budgets on QuALITY (Qwen3-1.7B, 4B, 8B, Llama-3.1-8B) and LongHealth (three Qwen3 models), the fold is within noise of AM's accuracy in 13 of 14 settings, with lower KL to the full cache than dropping the bias in all 12 where both ran.
Under the published per-head budgets, on Qwen3-4B, its Instruct-2507 variant and Llama-3.1-8B, it is within noise of AM in 9 of 10, no longer beats dropping the bias in KL, and the two each win once in accuracy.
\end{abstract}

%% file: sections/intro.tex
\section{Introduction}
\label{sec:intro}

A model answering questions about a long document holds a key and a value for every token in every layer.
For Qwen3-4B in fp16 that is 144\,KiB per token, and vLLM's KV pool on one T4 holds 4 requests over QuALITY prompts of 6,533 tokens on average, with 256 generated tokens each.
Attention Matching (AM)~\citep{zweiger2026am} shrinks this cache once the document is read.
Each head keeps a small subset of the keys and refits their values by least squares.
A kept key now stands in for many dropped ones, so AM also adds a scalar bias $\beta_j$ to its logit, $q\cdot k_j/\sqrt{d}+\beta_j$, fitted so that the kept keys carry the attention mass of the full context.
Later methods keep the bias, whether they learn part of the cache~\citep{shen2026arckv,chen2026latent} or apply AM to reasoning traces~\citep{liu2026tam} and agent turns~\citep{liu2026online}.

Serving systems have nowhere to put the bias.
FlashAttention~\citep{dao2023flashattention2} and vLLM's paged caches~\citep{kwon2023pagedattention} hold plain key and value tensors, PyTorch's flash backend rejects any additive mask, and prefix caches and KV transfer between engines move plain blocks.
The obstacle is availability, not speed.
A biased kernel costs little per call: AM reports a negligible change in attention runtime, ARC-KV measures its bias-aware path at 1.06 to 1.09 times the time of FlashAttention-2~\citep{shen2026arckv}, and our own timings agree (Section~\ref{sec:deploy}).
Yet every engine serving the cache needs this second, biased path, written and tuned for each GPU and cache layout; paged and prefix caches, KV transfer and FlashAttention itself have none, and papers that keep a per-entry logit term name this as their deployment gap~\citep{tian2025keepkv,cui2025asymkv,oneill2026still,chen2026latent}.
Two plain caches exist today.
An exact one adds a coordinate to every query and key, the rank-one case of FlashBias~\citep{wu2025flashbias}: it changes the model's query path, and vLLM would store its keys and values at head dimension 136 with FlashAttention or 256 with FlashInfer instead of 128 (Section~\ref{sec:related}).
The other drops $\beta$, as Still~\citep{oneill2026still} and QREAM~\citep{qream} do, and gives up the mass correction AM fits it for.

We keep the bias and move it into the keys.
If a kept key moves to $k_j'=k_j+\beta_j\sqrt{d}\,u$ for a direction $u$, its logit becomes $q\cdot k_j/\sqrt{d}+\beta_j\,(q\cdot u)$.
The bias adds $\beta_j$ for every query, the move adds $\beta_j(q\cdot u)$, and the two agree whenever $q\cdot u=1$.
Prior work~\citep{exactfinite} uses this identity for queries known in advance.
Future queries are not known, so we fit $u$ per head in closed form on the queries it produced while reading the context: $q\cdot u$ has mean 1 there, with, up to a small ridge, the smallest variance any such direction allows.
The folded and biased outputs then differ by at most $\tfrac32 D_V|q\cdot u-1|$, where $D_V$ is the largest distance between two attended values, and a value refit absorbs part of this error on those queries.
Dropping $\beta$ from AM's cache is the fold with $q\cdot u=0$, so when the attended biases vary little the fold's error is, to first order, $|q\cdot u-1|$ times the error of dropping $\beta$.
The fold gives a plain cache the size of AM's, at the model's own head dimension and with its query path unchanged, so with equal budgets per head any kernel that runs an uncompacted cache runs it (Section~\ref{sec:method-cost}).
The fold removes the obstacle the bias poses.
Per-head budgets pose a separate one, ragged per-head cache lengths, which AM's published schedule needs from the engine's memory manager whether or not $\beta$ is kept; we therefore serve equal budgets and measure per-head budgets for accuracy (Appendix~\ref{app:budget}).

We contribute:
\begin{itemize}
\item \emph{The fold.} One closed-form direction per head turns a biased cache into a plain one with the same size, head dimension and query path, with a bound on the output error in terms of $|q\cdot u-1|$ and a first-order comparison with dropping the bias (Section~\ref{sec:method}).
\item \emph{Accuracy.} With equal budgets per head, on QuALITY with four models and LongHealth with three, the folded cache is within noise of AM's accuracy in 13 of 14 settings. Its KL to the full cache is lower than dropping $\beta$ in all 12 settings where both ran, beyond noise in 10; dropping $\beta$ leaves more questions with KL above 1 in 11 of 12, and in accuracy the two differ beyond noise once, by 4.5 points in the fold's favor (Section~\ref{sec:downstream}, Tables~\ref{tab:downstream-quality} and~\ref{tab:downstream-longhealth}). Under the published per-head budgets it is within noise of AM in 9 of 10 settings and level with drop in KL (Appendix~\ref{app:budget}). Section~\ref{sec:downstream} and Appendix~\ref{app:llama-lh} diagnose the two losses.
\item \emph{Beyond AM.} The fold applies structurally to any per-entry logit term that does not depend on the query, which a census finds in AM's successors and in log-count merging (Table~\ref{tab:census}). It reproduces a capped log-count merged cache within noise on QuALITY and LongHealth and runs inside KV-Surgeon's own pipeline~\citep{wagner2026kvsurgeon} on two biased compactors with no detected loss (Section~\ref{sec:downstream}).
\item \emph{Serving.} Unmodified vLLM serves folded caches through a KV connector at the model's own head dimension. With equal budgets per head, answers match Hugging Face on all 51 Qwen3-4B questions, and greedy output matches on 180 of 200 LongBench samples with Qwen3-14B; capacity and decode-speed runs with random weights extend to Qwen3-32B (Section~\ref{sec:deploy}).
\end{itemize}

%% file: sections/background.tex
\section{Background: Attention-Matching Compaction}
\label{sec:background}

AM~\citep{zweiger2026am} treats every key-value head of every layer the same way, so we describe one head.
The context holds $n$ tokens with keys $K\in\mathbb{R}^{n\times d}$ and values $V\in\mathbb{R}^{n\times d}$, where $d$ is the head dimension and keys already carry their rotary position encoding.
For a query $q$, let $p_i(q)\propto\exp(q\cdot k_i/\sqrt{d})$ be its attention over the full context and $o(q)=\sum_i p_i(q)\,v_i$ its output.
AM fits the compact cache on a set of reference queries $Q\in\mathbb{R}^{M\times d}$, with rows $q_1,\dots,q_M$, that the model produces over the context.
Under grouped-query attention, $Q$ pools the queries of all query heads that share the key-value head.
In our benchmarks they come, as in AM, from a second prefill in which the model is asked to repeat the context, and in the layer study of Section~\ref{sec:why} they are the last queries of the original prefill.

AM keeps a set $S$ of $r\ll n$ keys with high attention over $Q$ (Section~\ref{sec:ablations} compares two ranking rules).
It gives each a weight $w_j$, so that the key counts as $w_j$ copies of itself in the softmax and stands in for dropped ones, and fits the weights by nonnegative least squares so that the reweighted keys carry the full attention mass of every reference query:
\begin{equation}
\min_{w}\;\sum_{m=1}^{M}\Big(\sum_{j\in S} w_j\,p_j(q_m)-1\Big)^2,
\qquad e^{-3}\le w_j\le e^{3}.
\label{eq:nnls}
\end{equation}
In the compact cache the weight becomes a bias $\beta_j=\log w_j\in[-3,3]$ added to the logit, $q\cdot k_j/\sqrt{d}+\beta_j$.
Last, AM refits the retained values by least squares,
\begin{equation}
V_S'=\arg\min_{X\in\mathbb{R}^{r\times d}}\;\big\|\operatorname{softmax}\!\big(QK_S^\top/\sqrt{d}+\mathbf{1}\beta^\top\big)\,X-O\big\|_F^2,
\label{eq:vfit}
\end{equation}
where $K_S$ stacks the retained keys, the softmax runs over each row, and row $m$ of $O$ is $o(q_m)$.
The compact cache is $(K_S,\beta,V_S')$.

%% file: sections/method.tex
\section{Folding the Bias into the Keys}
\label{sec:method}

\subsection{Goal}
\label{sec:method-goal}

We look for an ordinary key $k_j'$ that gives each retained key its AM logit,
\begin{equation}
\frac{q\cdot k_j'}{\sqrt{d}}\;\approx\;\frac{q\cdot k_j}{\sqrt{d}}+\beta_j,
\label{eq:goal}
\end{equation}
for the queries the head will see after compaction.
When $\beta_j\neq0$, no $k_j'$ satisfies \eqref{eq:goal} for every $q$, since the left side is linear in $q$ and the right side has a constant term.
For a finite set of queries known in advance the exact condition is $Q\,(k_j'-k_j)=\sqrt{d}\,\beta_j\mathbf{1}$~\citep{exactfinite}.
We know future queries only through the context queries, so we fit on those and measure how the fit carries over to later queries (Sections~\ref{sec:why} and~\ref{sec:downstream}).

\subsection{One shared direction per head}
\label{sec:method-u}

\begin{figure}[t]
\centering
\includegraphics[width=\linewidth]{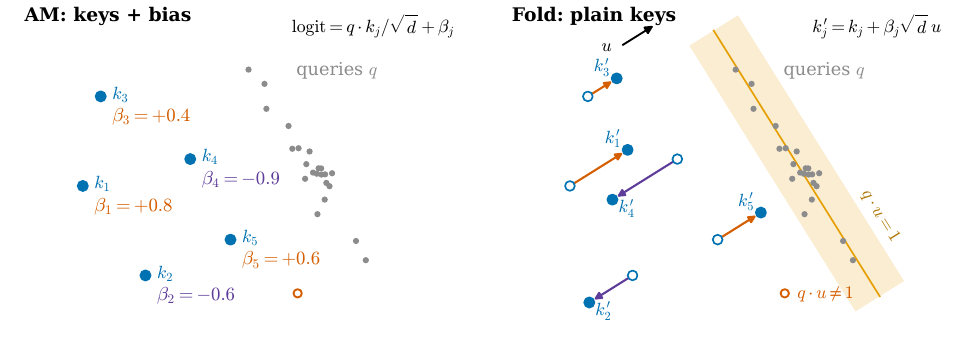}
\caption{The fold moves each retained key along one shared direction $u$ by $\beta_j\sqrt{d}$, so a query near the hyperplane $q\cdot u=1$ (shaded) sees the logit shift $\beta_j\,(q\cdot u)\approx\beta_j$. Left: AM, a scalar bias $\beta_j$ per retained key $k_j$. Right: the fold. For the orange open circle, a query far from the hyperplane, AM still adds $\beta_j$ while the fold scales it by $q\cdot u$.}
\label{fig:fold}
\end{figure}

Moving a key changes its logit by the query's projection on the move, so if every retained key of a head moves along one shared direction $u\in\mathbb{R}^d$ in proportion to its bias, each logit shifts by $\beta_j$ times the same scalar $q\cdot u$:
\begin{equation}
k_j'=k_j+\beta_j\sqrt{d}\,u
\qquad\Longrightarrow\qquad
\frac{q\cdot k_j'}{\sqrt{d}}=\frac{q\cdot k_j}{\sqrt{d}}+\beta_j\,(q\cdot u).
\label{eq:fold}
\end{equation}
The folded cache is AM with every bias of the head multiplied by $q\cdot u$ (Figure~\ref{fig:fold}): a query with $q\cdot u=1$ sees exactly the AM logits, and one with $q\cdot u=0$ sees the biases dropped.
Fitting each key's move separately by least squares on the same context queries gives $\beta_j\sqrt d$ times one shared direction, the pseudoinverse of the query matrix applied to $\mathbf 1$, so sharing $u$ costs nothing under this objective.
The objective asks that the head's queries lie near the hyperplane $q\cdot u=1$.

AM fits $\beta$ and the values on $Q$, and we fit $u$ on a set $Q_c\supseteq Q$ of context queries with rows $q_1,\dots,q_N$ (all original-prefill queries in Section~\ref{sec:why}, and $Q$ itself on the benchmarks).
With $\mu$ and $\Sigma$ their mean and covariance (normalized by $N$), the least-squares objective splits into a mean term, which puts the average query on the hyperplane, and a spread term, which keeps the queries close to it:
\begin{equation}
\frac1N\,\|Q_cu-\mathbf{1}\|^2=\frac1N\sum_{n=1}^{N}(q_n\cdot u-1)^2=(\mu^\top u-1)^2+u^\top\Sigma u .
\label{eq:split}
\end{equation}
Setting the gradient to zero gives $\Sigma u=(1-\mu^\top u)\,\mu$, so the minimizer points along $\Sigma^{-1}\mu$, and solving for its length gives $u_{\mathrm{LS}}=\Sigma^{-1}\mu/(1+s)$ with $s=\mu^\top\Sigma^{-1}\mu$.
Least squares trades mean for spread: on average over $Q_c$, $u_{\mathrm{LS}}$ applies only the fraction $s/(1+s)$ of each bias.
We keep its direction and rescale it so that the mean term of \eqref{eq:split} vanishes:
\begin{equation}
u=\frac{\Sigma^{-1}\mu}{\mu^\top\Sigma^{-1}\mu}.
\label{eq:u}
\end{equation}

This $u$ also minimizes $u^\top\Sigma u$ subject to $\mu^\top u=1$, so it has the smallest spread of $q\cdot u$ among directions with mean exactly $1$.
The factor $\Sigma^{-1}$ turns $u$ away from directions in which the queries scatter, toward the component they share.
On $Q_c$, $q\cdot u$ then has mean $1$ and (without the ridge) variance $1/s$, so its coefficient of variation is $s^{-1/2}$.
The query covariance can be ill-conditioned, so we use $\Sigma+\lambda I$ in place of $\Sigma$ in \eqref{eq:u}, with $\lambda=10^{-2}\operatorname{tr}(\Sigma)/d$, and the mean of $q\cdot u$ on $Q_c$ stays exactly $1$.

\subsection{Error bound}
\label{sec:method-bound}

Write $t(q)=q\cdot u-1$.
By \eqref{eq:fold}, the folded logits of a query are its AM logits plus $t(q)\,\beta_j$ on every retained key.
Entries without a bias, such as the question tokens, have $\beta_j=0$ in both caches.
If all attended keys had the same bias, this shift would raise every logit equally and the softmax would ignore it, so the output moves only when biases and values vary together.
We compare the two outputs with the same values $v_j$ in both caches.
Along the path $\ell(\tau)=\ell^{\mathrm{AM}}+\tau\,t(q)\,\beta$ for $\tau\in[0,1]$, the attention weights $p(\tau)=\operatorname{softmax}(\ell(\tau))$ give the output $o(\tau)=\sum_j p_j(\tau)\,v_j$, and the softmax derivative $\partial p_j/\partial\ell_i=p_j(\delta_{ij}-p_i)$ gives
\begin{equation}
\frac{do}{d\tau}=t(q)\sum_j p_j(\tau)\,(\beta_j-\bar\beta_\tau)\,v_j=t(q)\,\operatorname{Cov}_{p(\tau)}(\beta,v),
\label{eq:deriv}
\end{equation}
where $\bar\beta_\tau=\sum_j p_j(\tau)\beta_j$ and $\operatorname{Cov}_p(\beta,v)=\sum_j p_j(\beta_j-\bar\beta)(v_j-\bar v)$ is the covariance under the attention weights.
For a unit vector $w$, $w\cdot\operatorname{Cov}_p(\beta,v)=\operatorname{Cov}_p(\beta,w\cdot v)\le\operatorname{sd}_p(\beta)\,\operatorname{sd}_p(w\cdot v)$ by Cauchy--Schwarz.
Every $\beta_j$ lies in $[-3,3]$ and every $w\cdot v_j$ in an interval of length at most $D_V=\max_{i,j}\|v_i-v_j\|$, the largest distance between two values the query attends to, cached or not, so Popoviciu's inequality gives $\operatorname{sd}_p(\beta)\le3$ and $\operatorname{sd}_p(w\cdot v)\le D_V/2$ for any $p$.
Integrating \eqref{eq:deriv} over $\tau$ gives
\begin{equation}
\big\|o^{\mathrm{fold}}(q)-o^{\mathrm{AM}}(q)\big\|\;\le\;\tfrac32\,D_V\,\big|q\cdot u-1\big|.
\label{eq:bound}
\end{equation}

Over a set of queries on which $q\cdot u$ has mean $m$ and standard deviation $\sigma$, the mean of $|q\cdot u-1|$ is at most $\sqrt{(m-1)^2+\sigma^2}$.
On $Q_c$, $m=1$, so the mean error is at most $\tfrac32 D_V\,\sigma$, with $D_V$ taken as its largest value over these queries.
On later queries both terms can grow.

The constant is reached to first order in $t(q)$ when attention splits evenly between keys with $\beta=-3$ and $\beta=3$ whose values lie $D_V$ apart.
To first order the error is $t(q)\operatorname{Cov}_p(\beta,v)$, small whenever the attended keys have similar biases or similar values.
The bound holds for fixed values, which the refit of Section~\ref{sec:method-refit} changes after the fold.
It names two candidate causes of error on later queries: drift of the mean of $q\cdot u$, and its spread.

\paragraph{Fold against dropping the bias.}
Dropping $\beta$ while keeping AM's keys and values is the fold with $q\cdot u=0$, that is $t(q)=-1$.
To first order in $t(q)$ and in the spread of the attended biases, the fold's error is $|q\cdot u-1|$ times the error of dropping $\beta$ from AM's cache with the same values, not the \emph{drop} baseline of Section~\ref{sec:downstream}, which refits its values; Appendix~\ref{app:corollary} states this as a corollary with a bound on the remainder and checks the scaling on Llama-3.1-8B LongHealth.
The bound and the corollary use only that $\beta_j$ does not depend on the query, so they cover any static per-entry logit term; for a term with range $[b_{\min},b_{\max}]$, which includes the 0 of unbiased entries, the constant $\tfrac32$ of \eqref{eq:bound} becomes $(b_{\max}-b_{\min})/4$ by the same Popoviciu step.

\subsection{Value refit}
\label{sec:method-refit}

After the fold the logits on the reference queries differ from the AM logits by $t(q_m)\beta_j$, so we refit the values by \eqref{eq:vfit} on the folded keys without a bias:
\begin{equation}
V_S''=\arg\min_{X\in\mathbb{R}^{r\times d}}\;\big\|\operatorname{softmax}\!\big(QK_S'^{\top}/\sqrt{d}\big)\,X-O\big\|_F^2 .
\label{eq:vrefit}
\end{equation}
The refit absorbs part of the fold residual on the reference queries, as AM's value fit absorbs the residual of its key selection.
We round $K_S'$ to the cache precision (bf16) before the refit, so the values fit the keys the kernel will read (Algorithm~\ref{alg:fold}).

\begin{algorithm}[t]
\caption{Folding AM biases into the keys of one key-value head.}
\label{alg:fold}
\begin{algorithmic}[1]
\Require keys $K$, values $V$, reference queries $Q$, context queries $Q_c$, budget $r$
\State $S,\beta\gets$ top-$r$ keys by mean attention mass, NNLS fit \eqref{eq:nnls}, $\beta_j=\log w_j$ \Comment{our fit, Section~\ref{sec:ablations}}
\State $\mu,\Sigma\gets$ mean and covariance of the rows of $Q_c$
\State $a\gets(\Sigma+\lambda I)^{-1}\mu$ with $\lambda=10^{-2}\operatorname{tr}(\Sigma)/d$
\State $u\gets a/(\mu^\top a)$
\State $K_S'\gets K_S+\sqrt{d}\,\beta\,u^\top$, rounded to the cache precision \Comment{rank-one update}
\State $V_S''\gets$ least-squares solution of \eqref{eq:vrefit}
\State \Return $(K_S',V_S'')$ \Comment{a plain KV cache, no $\beta$}
\end{algorithmic}
\end{algorithm}

\subsection{Cost}
\label{sec:method-cost}

Per key-value head, the fold adds one covariance of the context queries, one $d\times d$ linear solve and a rank-one update of the $r$ retained keys, all small next to AM's NNLS and value fits, and its value refit replaces AM's at the same size.
The output is $r$ keys and $r$ values of dimension $d$ per head, the same tensors an uncompacted cache holds.
With equal budgets per head that is a plain cache of $r$ tokens.

%% file: sections/why.tex
\section{Where the Fold Tracks AM, Head by Head}
\label{sec:why}

By \eqref{eq:bound}, the fold's error over a set of queries grows with the spread $\sigma$ of $q\cdot u$ and with the drift of its mean $m$ away from $1$.
The fit sets $m=1$ on the queries it sees, and nothing holds it there thousands of tokens later.

\paragraph{Setup.}
We run Qwen3-1.7B, 4B and 8B on the first 32k tokens of \emph{Pride and Prejudice} (Project Gutenberg \#1342) and compact the first $n=4096$ tokens to $r=409$ keys per head (10\%).
AM (our fit, Section~\ref{sec:ablations}) and the folds' value refit \eqref{eq:vrefit} use the last 1024 context queries as $Q$ in place of the repeat prefill of Section~\ref{sec:downstream}, so every measured window starts a known distance $\Delta$ after the queries the fit saw.

We compare \emph{keep} (retained keys, original values, no bias), \emph{AM}, the \emph{base fold} with $u$ fitted on $Q$, and the \emph{default fold} with $u$ fitted on all 4096 context queries $Q_c$.
On every third layer (80 key-value heads for 1.7B, 96 for 4B and 8B) we measure $m$, $\sigma$ and the relative error $\|\hat o(q)-o(q)\|/\|o(q)\|$ on windows of 512 queries starting $\Delta$ tokens after the context.
All three measures pool the query heads that share a $u$.
A fold's excess is its error minus AM's. Near windows have $\Delta\le1$k, far ones $\Delta\ge4$k.
Errors are relative to attention renormalized over the 4096 compacted tokens (Section~\ref{sec:limits}).

\begin{figure}[t]
\centering
\includegraphics[width=\linewidth]{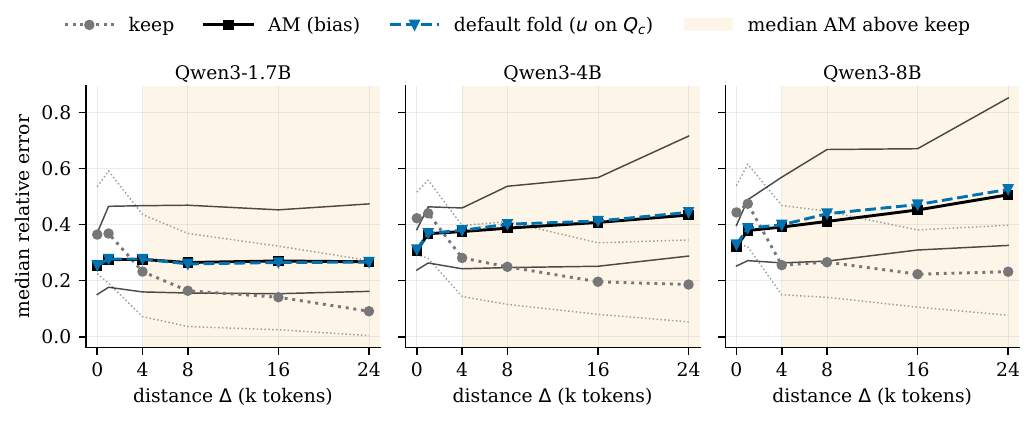}
\caption{On the median head the default fold stays within about 5\% of AM at every distance (the mean, worse far out, is in Appendix~\ref{app:far}). Median relative attention-output error over heads on \emph{Pride and Prejudice}, on 512-query windows starting $\Delta$ tokens after the compacted context. Thin lines: interquartile range over heads for keep (dotted) and AM (solid). Shaded: median AM error above median keep error.}
\label{fig:why-err}
\end{figure}

\begin{table}[t]
\centering
\caption{On \emph{Pride and Prejudice} the default fold tracks AM closely on the median head, and a tenth of the far (head, window) pairs carries most of its excess error. Ratio: median of $e_{\mathrm{fold}}/e_{\mathrm{AM}}$ over (head, window) pairs, pooled over near or far windows. Worse ${>}5\%$: share of far pairs with $e_{\mathrm{fold}}/e_{\mathrm{AM}}>1.05$. Worst 10\%: share of the total excess $\max(e_{\mathrm{fold}}-e_{\mathrm{AM}},0)$ in the worst tenth of far pairs. AM$<$keep: share of key-value heads on which AM's error is below keep's, in the window at that $\Delta$.}
\label{tab:why}
\small
\setlength{\tabcolsep}{3pt}
\begin{tabular}{lcccccc}
\toprule
 & \multicolumn{2}{c}{ratio} & worse ${>}5\%$ & worst 10\% & \multicolumn{2}{c}{AM$<$keep (\% heads)} \\
\cmidrule(lr){2-3}\cmidrule(lr){4-4}\cmidrule(lr){5-5}\cmidrule(lr){6-7}
model & near & far & far (\%) & far (\%) & $\Delta=0$ & $24$k \\
\midrule
Qwen3-1.7B & 1.005 & 0.998 & 11 & 90 & 81 & 18 \\
Qwen3-4B   & 1.011 & 1.021 & 35 & 92 & 85 & 17 \\
Qwen3-8B   & 1.011 & 1.022 & 40 & 96 & 82 & 17 \\
\bottomrule
\end{tabular}
\end{table}

\paragraph{Near the context.}
Near the context the default fold matches AM: at each near distance its median error ratio to AM is within 1.3\% of 1, and the median head keeps at least 97\% of AM's error reduction over keep.
Fitting $u$ on $Q_c$ rather than $Q$ is what gets it there, removing more than half of the base fold's excess (Appendix~\ref{app:far}).

\paragraph{Far queries.}
Far from the context the default fold still follows AM on the median head: pooled over far windows, the median ratio of their errors is within 2.2\% of 1 (Table~\ref{tab:why}).
The excess sits in a few heads: a tenth of far (head, window) pairs, mostly in layers 0 to 6, carries 90 to 96\% of it, and on those pairs the mean of $q\cdot u$ has drifted from 1 (Appendix~\ref{app:far}).
Here AM is fitted on the last 1024 context queries only, and from $\Delta=4$k its error exceeds keep's on the median head in every model (Figure~\ref{fig:why-err}, shaded), so the far windows measure how closely the fold tracks AM, not either method's absolute quality.

%% file: sections/downstream.tex
\section{Downstream Accuracy}
\label{sec:downstream}

On QuALITY with Qwen3-1.7B, 4B, 8B and Llama-3.1-8B-Instruct and on LongHealth with the three Qwen3 models, we detect no accuracy difference between the fold and AM in 13 of 14 settings (Tables~\ref{tab:downstream-quality} and~\ref{tab:downstream-longhealth}).
The exception is Qwen3-1.7B on LongHealth at 10\%.
In the other 13 every fold$-$AM interval lies between $-3.8$ and $+5.0$ points, so these benchmarks resolve only differences of about 4 to 5 points or more.

\paragraph{Setup.}
We use the first 30 articles of the QuALITY validation set~\citep{pang2022quality}, with 534 four-option questions over 2.8k to 8.1k tokens per article.
LongHealth~\citep{adams2024longhealth} has 20 patients, with 398 five-option questions over 10.0k to 13.2k tokens of clinical documents each.
Each context is prefilled once after a one-line instruction, in the model's chat template, with Qwen3's thinking mode off (Appendix~\ref{app:details}).
The reference queries $Q$ come from the repeat prefill of Section~\ref{sec:background}, at every second position.
They cover the whole document, so $Q_c=Q$.
Every key-value head keeps 10\% or 5\% of the context tokens.
The question and its options follow the compacted cache at their original positions, and the option letters' log-probabilities at the next token, renormalized over the letters, give the model's answer distribution.

We report accuracy and $\mathrm{KL}(\text{full}\,\|\,\cdot)$, the divergence of that distribution from the full cache's.
On Qwen3-8B QuALITY at 10\% the 5\% of questions with the largest KL carry about half of the mean, so KL differences mostly reflect how badly each cache fails on hard questions.

Intervals are 95\% paired bootstraps over contexts (2000 resamples).
Differences are computed per question, so they can differ by 0.1 from differences of rounded table entries.
With only 20 clusters on LongHealth, these intervals are if anything too narrow.
They also miss GPU nondeterminism: between repeat runs the fold$-$AM difference moved by up to 6 of 534 questions (1.1 points) in accuracy, 1.3 in agreement and 0.03 in KL (Appendix~\ref{app:rerun}).
We call a difference beyond noise only if its whole printed interval lies on one side of zero and clears it by more than these margins, and inconclusive otherwise.
The margins come from Qwen3 on QuALITY, and we apply them unchanged to LongHealth and Llama.

All four caches retain the same keys.
\emph{Keep} uses the original values and no bias, and \emph{drop} refits the values by \eqref{eq:vfit} with $\beta=0$.
\emph{AM} adds $\beta$ to the logits.
The \emph{fold} uses $u$ from \eqref{eq:u} over all coordinates and values refitted by \eqref{eq:vrefit}.
Each column uses whichever of the two AM fits of Section~\ref{sec:ablations} gives AM the lower KL, chosen on the test set in AM's favor: the AM-style fit for Qwen3-1.7B on QuALITY and our fit elsewhere; with equal budgets, Llama-3.1-8B ran only our fit.
The question attends to the bf16 cache through an explicit fp32 softmax.
On stock bf16 kernels the fold's top option changes on at most 2.1\% of questions (Appendix~\ref{app:stock}).

Our main tables give every head the same budget; from here on AM means our implementation.
With equal budgets our AM scores 63.7 and 60.5\% on Qwen3-4B QuALITY at 10\% and 5\%, against 71.3 and 70.1\% for the published AM~\citep{zweiger2026am}, which gives each head its own budget. With that schedule the AM-style fit reaches 67.8 and 65.2\%, about half-way from our AM to the published one (Appendix~\ref{app:budget}); the published protocol differs in other ways too (Appendix~\ref{app:protocol}).
Under the schedule (Appendix~\ref{app:budget}), in five model--benchmark pairs at two budgets each, the fold is level with AM in accuracy in 9 of 10 cells and loses 4.5 points on Llama-3.1-8B LongHealth at 5\% (Appendix~\ref{app:llama-lh}).
Its KL is above AM's and its agreement with the full cache below AM's in all 10 cells, beyond noise only in that Llama cell.

\begin{table}[t]
\centering
\caption{On QuALITY no fold$-$AM accuracy difference in these settings is beyond noise. Fold$-$AM and drop$-$fold: paired differences with their 95\% bootstrap intervals over the 30 articles below them, accuracy differences in points. Bold: beyond noise. KL$>$1: questions with KL above 1 under drop and under the fold. 534 questions. Each column's AM fit was chosen on the test set for AM's lower KL (Setup). Drop was not run with the AM-style fit on QuALITY, so the Qwen3-1.7B columns have none, and Table~\ref{tab:app-sweep} gives drop under our fit.}
\label{tab:downstream-quality}
\footnotesize
\setlength{\tabcolsep}{1.7pt}
\begin{tabular}{lcccccccc}
\toprule
 & \multicolumn{2}{c}{Qwen3-1.7B} & \multicolumn{2}{c}{Qwen3-4B} & \multicolumn{2}{c}{Qwen3-8B} & \multicolumn{2}{c}{Llama-3.1-8B} \\
\cmidrule(lr){2-3}\cmidrule(lr){4-5}\cmidrule(lr){6-7}\cmidrule(lr){8-9}
AM fit & \multicolumn{2}{c}{AM-style} & \multicolumn{2}{c}{ours} & \multicolumn{2}{c}{ours} & \multicolumn{2}{c}{ours} \\
budget & 10\% & 5\% & 10\% & 5\% & 10\% & 5\% & 10\% & 5\% \\
\midrule
full cache & \multicolumn{2}{c}{55.8} & \multicolumn{2}{c}{69.1} & \multicolumn{2}{c}{73.0} & \multicolumn{2}{c}{71.3} \\
\midrule
\multicolumn{9}{l}{\emph{accuracy (\%)}} \\
keep       & 45.5 & 42.3 & 62.2 & 58.8 & 70.2 & 62.5 & 65.7 & 62.4 \\
drop       & -- & -- & 63.3 & 57.9 & 68.4 & 61.2 & 67.4 & 61.4 \\
AM         & 45.3 & 42.3 & 63.7 & 60.5 & 68.4 & 63.3 & 66.1 & 64.0 \\
fold       & 46.8 & 43.4 & 62.7 & 59.2 & 69.1 & 63.3 & 67.6 & 63.1 \\
fold$-$AM & +1.5 & +1.1 & $-$0.9 & $-$1.3 & +0.7 & +0.0 & +1.5 & $-$0.9 \\
 & {\tiny [$-$0.8,+3.9]} & {\tiny [$-$1.1,+3.3]} & {\tiny [$-$2.4,+0.4]} & {\tiny [$-$3.1,+0.4]} & {\tiny [$-$1.4,+3.0]} & {\tiny [$-$2.2,+1.9]} & {\tiny [+0.0,+3.2]} & {\tiny [$-$3.2,+1.1]} \\
drop$-$fold & -- & -- & +0.6 & $-$1.3 & $-$0.7 & $-$2.1 & $-$0.2 & $-$1.7 \\
 & & & {\tiny [$-$2.1,+3.3]} & {\tiny [$-$4.8,+2.8]} & {\tiny [$-$3.3,+1.8]} & {\tiny [$-$5.7,+1.9]} & {\tiny [$-$2.2,+1.7]} & {\tiny [$-$5.7,+2.3]} \\
\midrule
\multicolumn{9}{l}{\emph{$\mathrm{KL}(\text{full}\,\|\,\cdot)$}} \\
keep       & 4.21 & 5.41 & 2.07 & 2.97 & 1.64 & 2.43 & 0.36 & 0.53 \\
drop       & -- & -- & 2.34 & 3.45 & 1.50 & 2.24 & 0.41 & 0.61 \\
AM         & 2.06 & 2.64 & 1.68 & 2.48 & 1.11 & 1.75 & 0.34 & 0.48 \\
fold       & 2.38 & 2.97 & 1.68 & 2.50 & 0.90 & 1.24 & 0.34 & 0.47 \\
fold$-$AM & \textbf{+0.32} & \textbf{+0.33} & +0.00 & +0.02 & \textbf{$-$0.21} & \textbf{$-$0.51} & +0.00 & $-$0.01 \\
 & {\tiny [+0.11,+0.55]} & {\tiny [+0.17,+0.52]} & {\tiny [$-$0.16,+0.15]} & {\tiny [$-$0.16,+0.20]} & {\tiny [$-$0.39,$-$0.05]} & {\tiny [$-$0.64,$-$0.38]} & {\tiny [$-$0.01,+0.01]} & {\tiny [$-$0.03,+0.01]} \\
drop$-$fold & -- & -- & \textbf{+0.66} & \textbf{+0.95} & \textbf{+0.60} & \textbf{+1.00} & +0.07 & \textbf{+0.14} \\
 & & & {\tiny [+0.43,+0.92]} & {\tiny [+0.67,+1.25]} & {\tiny [+0.40,+0.81]} & {\tiny [+0.73,+1.29]} & {\tiny [+0.03,+0.12]} & {\tiny [+0.08,+0.21]} \\
KL$>$1: drop/fold & -- & -- & 126/107 & 168/141 & 117/102 & 170/144 & 55/49 & 103/74 \\
\bottomrule
\end{tabular}
\end{table}

\begin{table}[t]
\centering
\caption{On LongHealth the fold keeps AM's accuracy gain over keep on Qwen3-4B and 8B and loses it on Qwen3-1.7B at 10\%. Our AM fit throughout. Layout as in Table~\ref{tab:downstream-quality}, with intervals over the 20 patients (398 questions).}
\label{tab:downstream-longhealth}
\small
\setlength{\tabcolsep}{4pt}
\begin{tabular}{lcccccc}
\toprule
 & \multicolumn{2}{c}{Qwen3-1.7B} & \multicolumn{2}{c}{Qwen3-4B} & \multicolumn{2}{c}{Qwen3-8B} \\
\cmidrule(lr){2-3}\cmidrule(lr){4-5}\cmidrule(lr){6-7}
budget & 10\% & 5\% & 10\% & 5\% & 10\% & 5\% \\
\midrule
full cache & \multicolumn{2}{c}{68.3} & \multicolumn{2}{c}{77.4} & \multicolumn{2}{c}{80.4} \\
\midrule
\multicolumn{7}{l}{\emph{accuracy (\%)}} \\
keep       & 36.2 & 30.4 & 46.7 & 39.7 & 44.7 & 37.2 \\
drop       & 38.9 & 33.9 & 54.3 & 45.2 & 51.3 & 42.5 \\
AM         & 39.9 & 31.9 & 54.5 & 43.5 & 55.8 & 45.0 \\
fold       & 35.4 & 33.4 & 55.0 & 46.0 & 55.8 & 44.2 \\
fold$-$AM & \textbf{$-$4.5} & +1.5 & +0.5 & +2.5 & +0.0 & $-$0.8 \\
 & {\tiny [$-$7.6, $-$1.8]} & {\tiny [$-$2.0, +5.0]} & {\tiny [$-$1.0, +2.3]} & {\tiny [+0.5, +4.5]} & {\tiny [$-$2.8, +2.8]} & {\tiny [$-$3.8, +2.0]} \\
drop$-$fold & +3.5 & +0.5 & $-$0.8 & $-$0.8 & \textbf{$-$4.5} & $-$1.8 \\
 & {\tiny [+0.5, +6.5]} & {\tiny [$-$3.5, +4.5]} & {\tiny [$-$3.5, +2.3]} & {\tiny [$-$3.8, +2.3]} & {\tiny [$-$7.3, $-$1.3]} & {\tiny [$-$5.8, +2.3]} \\
\midrule
\multicolumn{7}{l}{\emph{$\mathrm{KL}(\text{full}\,\|\,\cdot)$}} \\
keep       & 4.04 & 4.79 & 4.92 & 5.57 & 3.00 & 3.85 \\
drop       & 3.59 & 3.59 & 3.90 & 5.32 & 2.41 & 3.08 \\
AM         & 3.05 & 3.11 & 3.47 & 4.66 & 1.92 & 2.63 \\
fold       & 3.07 & 3.20 & 3.33 & 4.41 & 1.81 & 2.27 \\
fold$-$AM & +0.02 & +0.09 & $-$0.14 & $-$0.25 & $-$0.11 & \textbf{$-$0.36} \\
 & {\tiny [$-$0.27, +0.30]} & {\tiny [$-$0.25, +0.42]} & {\tiny [$-$0.32, +0.03]} & {\tiny [$-$0.49, +0.00]} & {\tiny [$-$0.24, +0.01]} & {\tiny [$-$0.56, $-$0.15]} \\
drop$-$fold & \textbf{+0.52} & +0.39 & \textbf{+0.57} & \textbf{+0.91} & \textbf{+0.60} & \textbf{+0.81} \\
 & {\tiny [+0.26, +0.78]} & {\tiny [$-$0.13, +0.91]} & {\tiny [+0.16, +0.98]} & {\tiny [+0.44, +1.42]} & {\tiny [+0.39, +0.81]} & {\tiny [+0.46, +1.17]} \\
KL$>$1: drop/fold & 195/200 & 223/218 & 148/142 & 194/181 & 152/130 & 195/184 \\
\bottomrule
\end{tabular}
\end{table}

\paragraph{QuALITY.}
On QuALITY the bias buys little accuracy at these budgets: every AM$-$keep accuracy interval contains zero.
KL separates the caches more. AM's gain over keep is clear on Qwen3 and small on Llama, where even keep stays close to the full cache. On Qwen3-8B the fold's KL is below AM's, beyond noise; on Qwen3-1.7B it keeps 85\% and 88\% of AM's KL reduction over keep but stays above AM, beyond noise.

\paragraph{LongHealth.}
AM's gain over keep is clear in three columns, Qwen3-4B at 10\% and Qwen3-8B at both budgets, and the fold keeps it: on Qwen3-4B and 8B it beats keep by 6.3 to 11.1 points in all four columns and is level with AM (Table~\ref{tab:downstream-longhealth}).
Its KL is lower than AM's in all four columns, beyond noise on 8B at 5\%, and on 4B at 5\% it agrees with the full cache's top option more often than AM, beyond noise (Table~\ref{tab:app-agree}).
The fold fails in one cell, Qwen3-1.7B at 10\%, where AM's gain over keep is marginal: the fold gives it up, landing 4.5 points below AM and no better than keep.
Under the AM-style fit the same cell is 3.0 points below AM, short of our margin (Table~\ref{tab:app-lh-recipe}).

\paragraph{Against dropping the bias.}
Drop is the plain cache an engine could serve today; the fold is the plain cache closest to AM's.
Without $\beta$ the kept keys carry only their own attention mass, which underestimates the document's share of attention against the tokens that follow it~\citep{zweiger2026am}.
With equal budgets per head the fold's KL is lower than drop's in all 12 columns, beyond noise in 10, and drop has more questions with KL above 1 in 11 of the 12 (Tables~\ref{tab:downstream-quality} and~\ref{tab:downstream-longhealth}).
The order holds in the other settings of Table~\ref{tab:app-sweep} and flips once, within noise, in Table~\ref{tab:app-lh-recipe}.
Drop's agreement with the full cache's top option is also lower in 11 of the 12 (Table~\ref{tab:app-agree}).
In accuracy the two differ beyond noise once, by 4.5 points in the fold's favor on Qwen3-8B LongHealth at 10\%; on LongHealth drop recovers 59 to 97\% of AM's gain over keep in four settings and all of it in the other two.
Under the published per-head budgets the KL advantage disappears: no drop$-$fold KL difference is beyond noise, and on Llama-3.1-8B LongHealth at 5\%, the cell where the fold loses to AM, drop is 7.8 points more accurate than the fold, beyond noise, while on Qwen3-4B LongHealth at 5\% the fold is 3.5 points more accurate than drop (Appendices~\ref{app:budget} and~\ref{app:llama-lh}).

\paragraph{Beyond AM.}
The fold applies structurally to any static logit term (Table~\ref{tab:census}).
We test it on log-count merging, where a kept key absorbs evicted keys whose post-RoPE cosine with it is at least 0.9, up to 16 per cluster, and its logit gains $\beta=\log n$ for a cluster of $n$ keys~\citep{bolya2022tome}, with Qwen3-4B under the published per-head budgets.
Folding the count into the keys reproduces the biased merged cache within noise on both benchmarks, with every accuracy difference within 0.6 points (Appendix~\ref{app:beyond}).
We also fold the fitted biases of two compactors inside the code and harness of KV-Surgeon~\citep{wagner2026kvsurgeon}, scoring answers generated with thinking on, with Qwen3-4B on 10 QuALITY articles at 5\%.
The fold runs inside this third-party pipeline with no detected loss: it is 3.3 points above KV-Surgeon-fast and 2.8 below its KVMerger variant, neither beyond noise.

%% file: sections/deploy.tex
\section{Serving Folded Caches}
\label{sec:deploy}

\paragraph{Answers in vLLM.}
We served folded caches on an unmodified vLLM~0.18.1 with its Triton attention backend, in fp16 on one Tesla T4, with Qwen3-4B and prefix caching off (Table~\ref{tab:serve}).
The caches are folds of the AM-style fit with equal budgets per head on the first 3 QuALITY articles (51 questions), with the prompt template left uncompacted, so a request holds 781 and 462 prompt tokens on average at 10\% and 5\%, against 6,533 for the full prompt.
A KV connector, loaded by module path with no change to vLLM's source, writes the folded keys and values into vLLM's paged blocks, and vLLM prefills only the question (Appendix~\ref{app:inject}).
The folded caches give the same answer letter as Hugging Face on all 51 questions at both budgets.
Their 128 greedy tokens match Hugging Face's decoding from the same cache exactly on 42 and 44 of the 51 questions; the full prompt itself matches on only 46, so part of the mismatch is fp16 differences between the two implementations (Appendix~\ref{app:inject}).

\paragraph{Answers from a 14B model.}
We also served folded caches of Qwen3-14B with its real weights, in bf16 on one A100-80GB with vLLM's default FlashAttention backend, on four English question-answering tasks of LongBench~\citep{bai2024longbench}: Qasper, MultiFieldQA-en, HotpotQA and 2WikiMQA (Table~\ref{tab:lb14}).
We use the benchmark's prompts, answer lengths and F1 score, with thinking off, on the first 50 samples of each task whose prompt and answer fit in 16,384 tokens; the full prompts average 8,187 tokens.
The caches are folds of the AM-style fit at 10\% with equal budgets per head, compacting only the context, so an injected request holds 905 prompt tokens on average.
Served by vLLM through the same connector, the folded caches give the same greedy output as Hugging Face decoding from the same cache on 180 of the 200 samples and the same first token on 197, with a mean F1 of 29.2 against 28.7 in Hugging Face.
In Hugging Face the fold's mean F1 is 2.0 points below AM's, with an interval that contains zero.
Drop's mean F1 is 30.1 against the fold's 28.7, and dropping $\beta$ costs AM only 0.7 points, so these tasks test serving more than the bias.

\paragraph{Capacity and speed.}
Compaction provides the gain over the full cache, and the fold lets the compacted cache run on this stock engine at the model's own head dimension.
With 256 generated tokens per request, the T4's KV pool holds 4 full requests, 31 at 10\% and 44 at 5\%, and peak decode rises from 29 to 200 and 297 tokens/s.
Plain prompts of the same lengths decode at 207 and 292 tokens/s at their largest batch, and at equal batches from 4 to 32 the folded caches decode at 97 to 98\% of their rate (88\% at batch 1).

\paragraph{Larger models.}
We repeated the capacity and speed measurements for Qwen3-14B and Qwen3-32B on one A100-80GB in bf16, with vLLM's default FlashAttention backend (Table~\ref{tab:serve-big}).
Capacity and decode speed depend only on the model's shapes and the cache lengths, so these engines load random weights of the real architectures and the connector injects random tensors of the fold's lengths; these runs test no answers.
To lengthen the context we also repeat each article 2 times (32B) or 4 times (14B).
On the 32B model one A100 holds 7 full requests of 6.5k tokens on average, or 45 at 10\%, and peak decode rises from 135 to 872 tokens/s; at equal batch sizes the injected caches decode at 99 to 101\% of the rate of plain prompts of the same length on both models.
An exact plain cache needs head dimension 129 and an extra coordinate on every query (Section~\ref{sec:related}).
vLLM~0.18.1 would store it at 136 with this FlashAttention backend, where by the pool's block count the 32B model would hold about 43 instead of 45 requests at 10\%, and at 256 with FlashInfer.
The fold keeps head dimension 128 and the query path unchanged.

\paragraph{Kernel cost of the bias.}
The fold buys availability: it runs where no biased kernel exists, and paged and prefix caches, KV transfer and FlashAttention have no bias input at all.
Where a biased kernel does exist it costs little, 1.02 to 1.29 times as long as the fastest plain one per single-sequence decode call and up to 1.90 times at batch 8.
We time one attention layer at each model's head shapes on an A100, with plain and biased PyTorch SDPA and FlexAttention paths captured in CUDA graphs as an engine replays its decode steps (Appendix~\ref{app:eager}).

\begin{table}[t]
\centering
\small
\caption{A folded Qwen3-4B cache served by unmodified vLLM~0.18.1 (Triton backend, fp16, one Tesla T4, 51 QuALITY questions, equal budgets per head). Capacity: concurrent requests that fit the KV pool with 256 generated tokens each. Decode: tokens/s at the largest batch that fits. Agreement with Hugging Face (HF) on the same cache: answer letter, and exact match of 128 greedy tokens. Correct letter: 51 questions are too few to measure accuracy, which Section~\ref{sec:downstream} does.}
\label{tab:serve}
\setlength{\tabcolsep}{4pt}
\begin{tabular}{lrrr}
\toprule
 & full & fold 10\% & fold 5\% \\
\midrule
prompt tokens (mean) & 6,533 & 781 & 462 \\
KV per token (KiB) & 144 & 144 & 144 \\
capacity & 4 & 31 & 44 \\
decode (tokens/s) & 29 & 200 & 297 \\
\midrule
same letter as HF & 51/51 & 51/51 & 51/51 \\
same 128 tokens as HF & 46/51 & 42/51 & 44/51 \\
correct letter & 35/51 & 25/51 & 23/51 \\
\bottomrule
\end{tabular}
\end{table}

\begin{table}[t]
\centering
\small
\caption{Qwen3-14B with real weights on four LongBench tasks at 10\%: the folded cache served by unmodified vLLM~0.18.1 scores as it does in Hugging Face (HF), and its mean is within noise of AM's. F1 $\times100$ with the official scorer, 50 samples per task, AM-style fit, equal budgets per head; bf16, one A100-80GB. Fold$-$AM: paired difference in the HF runs with its 95\% bootstrap interval over samples.}
\label{tab:lb14}
\setlength{\tabcolsep}{4pt}
\begin{tabular}{lrrrrr}
\toprule
 & Qasper & MF-en & HotpotQA & 2WikiMQA & mean \\
\midrule
full cache (HF) & 43.8 & 47.5 & 60.5 & 41.7 & 48.4 \\
AM (HF) & 24.1 & 30.7 & 37.8 & 30.4 & 30.8 \\
drop (HF) & 20.0 & 29.8 & 37.4 & 33.1 & 30.1 \\
fold (HF) & 28.8 & 29.9 & 29.1 & 27.0 & 28.7 \\
fold (vLLM) & 28.0 & 30.2 & 28.9 & 29.5 & 29.2 \\
\midrule
fold$-$AM & $+4.7$ & $-0.8$ & $-8.8$ & $-3.3$ & $-2.0$ \\
 & \scriptsize$[+0.7,+10.0]$ & \scriptsize$[-4.5,+2.4]$ & \scriptsize$[-19.3,+1.0]$ & \scriptsize$[-9.9,+1.8]$ & \scriptsize$[-5.3,+1.1]$ \\
\bottomrule
\end{tabular}
\end{table}

\begin{table}[t]
\centering
\small
\caption{Folded caches of larger models served by unmodified vLLM~0.18.1 (default FlashAttention backend, bf16, one A100-80GB, random weights of the real architectures, 51 QuALITY questions). Capacity: concurrent requests that fit the KV pool with 256 generated tokens each. Decode: the highest rate measured, in tokens/s, at the batch in parentheses. At 5\% the 14B model reaches vLLM's limit of 256 concurrent sequences, and at 10\% its pool holds 255, one short of it. The $\times2$ and $\times4$ rows repeat each article to lengthen the context. 14B peak rates are measured at batch $\le128$ and are lower bounds. Fold columns at the native length are injected caches; at the longer lengths they are plain prompts of the fold's length, which decode at the same rate.}
\label{tab:serve-big}
\setlength{\tabcolsep}{4pt}
\begin{tabular}{llrrrrrr}
\toprule
 & & \multicolumn{3}{c}{capacity} & \multicolumn{3}{c}{decode (tokens/s)} \\
\cmidrule(lr){3-5}\cmidrule(lr){6-8}
model & context (max) & full & 10\% & 5\% & full & 10\% & 5\% \\
\midrule
Qwen3-32B & 6,969 & 7 & 45 & 64 & 135 (7) & 872 (45) & 1,232 (64) \\
Qwen3-32B & 13,754 ($\times2$) & 3 & 28 & 45 & 59 (3) & 534 (28) & 864 (45) \\
Qwen3-14B & 6,969 & 39 & 255 & 256 & 766 (39) & 3,442 (128) & 3,831 (128) \\
Qwen3-14B & 27,324 ($\times4$) & 10 & 90 & 158 & 196 (8) & 1,000 (32) & 1,182 (32) \\
\bottomrule
\end{tabular}
\end{table}

%% file: sections/ablations.tex
\section{Ablations}
\label{sec:ablations}

Both AM fits solve the NNLS of \eqref{eq:nnls} by projected gradient.
The AM-style fit follows the settings \citet{zweiger2026am} give for their highest-attention-keys method, and the two differ in key ranking, starting point and number of steps (Appendices~\ref{app:details} and~\ref{app:protocol}).
The one that gives AM the lower KL is, on QuALITY, the AM-style fit on Qwen3-1.7B and ours on 4B and 8B, and on LongHealth ours on all three (Appendix~\ref{app:fits}).

We change one choice of the fold at a time on QuALITY, with Qwen3-1.7B, 4B and 8B at 10\% and 5\% (Appendix~\ref{app:abl}).
On 4B and 8B the value refit lowers the fold's KL by 0.36 to 0.67, more than it lowers AM's, and the covariance term in $u$ also helps: the mean direction $u=\mu/\|\mu\|^2$ raises the fold's KL in all four columns, beyond noise in two.
On 1.7B under our fit both effects reverse, where that fit is weak and AM is less accurate than keep at every budget; under the AM-style fit the refit lowers KL there too.

%% file: sections/related.tex
\section{Related Work}
\label{sec:related}

\paragraph{Compaction with a per-key bias.}
Methods built on AM~\citep{zweiger2026am} keep the bias.
ARC-KV~\citep{shen2026arckv} learns the key selection and still fits a mass bias, TAM~\citep{liu2026tam} splits the budget across reasoning segments before the same fits, and Latent Compaction~\citep{chen2026latent} trains biased cache entries at test time.
KV-Surgeon~\citep{wagner2026kvsurgeon} keeps original keys with a per-key bias in AM's $[-3,3]$ box, serves it through an additive attention mask and evaluates on QuALITY and LongHealth with Qwen3-4B; we fold its caches in its own code (Section~\ref{sec:downstream}).
\citet{liu2026online} carry AM, bias included, over to online compaction of agent turns.
Running such a cache means leaving fast kernels or dropping the bias: Latent Compaction computes its biased attention with explicit PyTorch operations, and the Still method~\citep{oneill2026still} sets $\beta=0$ in every reported result because serving the bias would need kernel support.

\paragraph{Which methods carry a static logit term.}
The fold needs only that each cache entry's logit term be the same for every query, and Table~\ref{tab:census} sorts compaction and merging methods by this property.
Besides AM and its successors, token- and KV-merging methods that add the log of a merged entry's size or mass carry such a term~\citep{bolya2022tome,wu2026semanticache,tian2025keepkv,liu2025zsmerge,cui2025asymkv,bouyahiaoui2026selkv}, but it is not clipped to $[-3,3]$: SelKV reports counts near 200 at 75\% compression, a log-count above 5, and the fold's error grows with the spread of the attended terms (Section~\ref{sec:method-bound}).
ResKV~\citep{zhan2026reskv} and MomentKV~\citep{li2026momentkv} add a query-dependent correction to a static log-count, and only the count folds.
ALiBi's positional bias is static per key too, but it is unbounded, and its exact rank-one augmentation is FlashBias's own example~\citep{wu2025flashbias}.
\begin{table}[t]
\centering
\caption{Logit terms of compaction and merging methods. Static: one scalar per cache entry, the same for every query, which the fold handles. Fold applies: structurally, since the fold has been tested only on AM, one capped log-count merge and two of KV-Surgeon's compactors.}
\label{tab:census}
\footnotesize
\setlength{\tabcolsep}{4pt}
\begin{tabular}{@{}>{\raggedright\arraybackslash}p{0.40\linewidth}lc@{}}
\toprule
method & per-entry logit term & fold applies \\
\midrule
AM~\citep{zweiger2026am} & $\beta=\log w$, NNLS fit, $|\beta|\le3$ & yes (this paper) \\
KV-Surgeon, ARC-KV, TAM, online AM~\citep{wagner2026kvsurgeon,shen2026arckv,liu2026tam,liu2026online} & fitted $\log w$ & yes \\
Latent Compaction~\citep{chen2026latent} & learned per-slot bias & yes, bias only$^{a}$ \\
Still~\citep{oneill2026still} & optional $\beta$, off in all results & yes \\
\midrule
ToMe, SemantiCache, KeepKV, ZSMerge, AsymKV~\citep{bolya2022tome,wu2026semanticache,tian2025keepkv,liu2025zsmerge,cui2025asymkv} & $\log$ count or size, $\alpha$-scaled in ZSMerge & yes \\
SelKV~\citep{bouyahiaoui2026selkv} & $0.5\log$ attention mass & yes \\
ResKV, MomentKV~\citep{zhan2026reskv,li2026momentkv} & $\log$ count $+$ query-dependent term & count term only \\
\midrule
KVMerger, D2O, MiniCache~\citep{wang2024kvmerger,wan2024d2o,liu2024minicache} & none & nothing to fold \\
ALiBi & $m\cdot j$, unbounded & exact augmentation instead \\
\bottomrule
\end{tabular}

\smallskip
{\footnotesize $^{a}$Latent Compaction also needs a ragged-slot mask and a log-sum-exp output, so the fold alone does not make its cache plain.}
\end{table}

\paragraph{Serving an attention bias.}
AM notes that PyTorch SDPA and FlexAttention accept an additive bias, and FlashInfer compiles custom attention variants, including logit transforms, just in time~\citep{ye2025flashinfer}.
FlashBias~\citep{wu2025flashbias} computes low-rank biases inside FlashAttention by appending factor dimensions to queries and keys, so its cache is no longer a plain one.
A per-key bias is its rank-one case, which gives an exact cache with plain tensors: give every query a constant extra coordinate and every kept key the matching $\beta_j\sqrt{d}$.
Head dimension 128 then becomes 129: in vLLM~0.18.1 the FlashAttention backend accepts multiples of 8, so the cache would be stored at 136, 6\% more KV memory than the fold's, and FlashAttention-2 runs it on its kernel for 192; the Triton backend stores 129 but computes at 256, the next power of two; and FlashInfer supports only 64, 128 and 256, so there it would be stored at 256.
The engine must also append the constant to every query, a change to the model's query path.
The fold keeps $d$ and the query path unchanged but is exact only on the hyperplane $q\cdot u=1$.

\paragraph{Dropping the bias and merging keys.}
QREAM~\citep{qream} argues that per-key biases are incompatible with standard KV caches and removes them, matching the normalized attention output directly and solving only for the values.
\emph{On the Limits of KV Cache Merging}~\citep{limits} proves that a value-only merge is exact for every future query only when evicting the token would have lost nothing.
It also finds that compensated merges which rewrite retained keys lose accuracy, and in its clearest case the loss goes away when the keys stay in place.
The fold rewrites keys too, with an error bounded by a multiple of $|q\cdot u-1|$ (Section~\ref{sec:method-bound}).

\paragraph{Finite-query compilation.}
\emph{Exact Finite-Query KV Cache Compilation}~\citep{exactfinite} gives the exact condition of Section~\ref{sec:method-goal} and uses it to certify caches on queries known in advance; we fit $u$ to carry over to unseen queries.
It also shows that duplicating a key $n$ times adds $\log n$ to its logit, which gives an exact plain cache for $\beta_j>0$ but stores up to $e^{3}\approx20$ copies of a key and cannot express $\beta_j<0$.
CertKV~\citep{yang2026certkv} reserves one tail-summary slot per head; the fold adds no slot.

\paragraph{Eviction and quantization of plain caches.}
Eviction by accumulated attention, an observation window or position~\citep{zhang2023h2o,li2024snapkv,xiao2024streamingllm} and 2-bit quantization~\citep{liu2024kivi} leave caches that standard kernels run.

%% file: sections/limits.tex
\section{Limitations}
\label{sec:limits}

\paragraph{What the benchmarks resolve.}
Our intervals can show the fold keeping AM's accuracy gain only where AM beats keep.
On QuALITY no AM$-$keep interval excludes zero, so those columns show only that the fold does no detectable harm; only three main LongHealth settings, on 20 patients, show a clear AM gain, and none of the six LongHealth cells under per-head budgets does.
Answers are scored by the log-probabilities of the options, so the fold's lower KL than drop's is not tied to generated answers or to later turns of a conversation; the only generated answers come from the KV-Surgeon runs and the LongBench serving run.
Under the published per-head budgets no drop$-$fold KL difference is beyond noise, and in the Llama cell where the fold loses to AM drop is more accurate than the fold (Appendix~\ref{app:budget}), so the fold's KL advantage over drop is shown only with equal budgets.

\paragraph{Where the fold loses to AM.}
The two losses are analysed in Section~\ref{sec:downstream} and Appendix~\ref{app:llama-lh}.
On Llama-3.1-8B LongHealth at 5\% under per-head budgets the fold's KL is also higher on 19 of the 20 patients.
On five patients the loss appears only with large biases and the value refit, and self-study queries or a $\pm1$ box on $\beta$ remove it; the $\pm1$ box also removes it on all 20.
The $\pm1$ box was chosen on this failure, is untested in other settings and is not used in any main table, which all keep $\pm3$.
The sign of the fold$-$AM KL difference also depends on the model.

\paragraph{Queries far from the context.}
Far from the context AM and the fold both lose to keep, and a few early-layer heads lift the fold's mean error above AM's (Appendix~\ref{app:far}).
The layer study compacts a single 4096-token prefix of one text, and its errors are relative to attention renormalized over the compacted span, which overstates the effect on the layer output by a factor that grows with distance and that we did not log.
Our QA benchmarks ask their questions right after the document, so they do not probe this regime; a multi-turn task would.

\paragraph{Serving.}
We checked vLLM's answers only with equal budgets per head, on Qwen3-4B and Qwen3-14B; the capacity and speed runs of Table~\ref{tab:serve-big}, on 14B and 32B, use random weights and test no answers.
We did not serve per-head budgets; our per-head-budget accuracy runs pad ragged heads and mask the padding, so they do not run as a compact plain cache.
The kernel timings cover one attention layer, one run per model on one GPU.

\paragraph{Scope.}
We tested self-study queries only on Qwen3-1.7B and in the Llama diagnosis, and did not test compacting a cache that was already compacted and folded, as online compaction would.
Outside AM we tested one log-count merge, whose cap keeps $\beta\le2.77$, and two of KV-Surgeon's compactors on 10 articles; the other rows of Table~\ref{tab:census} are structural.

%% file: sections/conclusion.tex
\section{Conclusion}
\label{sec:conclusion}

Moving each kept key along one fitted direction per head turns an AM cache into a plain KV cache that runs on standard attention kernels, with an output error bounded by a multiple of $|q\cdot u-1|$.
The folded cache is within noise of AM's accuracy in 13 of 14 settings with equal budgets per head and in 9 of 10 under the published per-head budgets, at a resolution of about 5 points, and with equal budgets per head unmodified vLLM serves it with its stock kernels.
Unlike an exact plain cache, it needs no extra head coordinate and no change to the query path.
It applies, structurally, to any static per-entry logit term; we confirm this on one capped log-count merge and two of KV-Surgeon's compactors, and its error grows with the spread of the term.
A next step is to test whether query drift, measured at compaction time, flags the heads that should keep their bias.

%% file: sections/appendix.tex
\section{Additional Results}
\label{sec:appendix}

\subsection{Protocol details}
\label{app:details}
Prompts use each model's chat template, with \texttt{enable\_thinking=False} for Qwen3.
The QuALITY prompt is ``Read the story, then answer the question.'' followed by the article, and the reference prefill appends ``Repeat the story above word for word.'' and the article again.
LongHealth does the same with the patient's documents, joined by the benchmark's own separator, and we drop the two of its 400 questions whose correct answer appears under two letters.
The question follows as ``Question:'', the lettered options and ``Answer with the letter only.'', and we read the option letters' log-probabilities at the first token of the reply.
Our fit compacts the whole context, instruction included; the AM-style fit leaves the template and instruction tokens before the text uncompacted and keeps $r$ of the text tokens.
All value fits are solved in double precision, by plain least squares without a ridge.
Both NNLS fits take projected-gradient steps of size $1/\Lambda$, where $\Lambda$ estimates $\|A\|_2^2$ for the NNLS matrix $A$, and clip every $w_j$ to $[e^{-3},e^{3}]$.
Our fit ranks keys by their mean attention over $Q$, starts each kept key with the attention mass of the dropped keys closest to it in cosine similarity, and runs 300 steps.
The AM-style fit uses root-mean-square ranking, a clamped least-squares start and two steps.

\subsection{The AM baseline against the published protocol}
\label{app:protocol}
The AM-style fit is our approximation of the highest-attention-keys method of \citet{zweiger2026am}, not a faithful reproduction.
On Qwen3-4B QuALITY it scores 61.6 and 53.9\% at 10\% and 5\%, against 63.7 and 60.5\% for our fit and 71.3 and 70.1\% for the published AM.
Their protocol scores answers generated by decoding from the compacted cache, over the first 50 QuALITY articles (894 questions).
It fits on up to 50,000 reference queries per head from self-study and the repeat prefill.
We read the option letters' log-probabilities over 30 articles (534 questions) and fit on repeat-prefill queries only, except in the self-study rows of Table~\ref{tab:abl-recipe}.
LongHealth is not shared: they run Qwen3-4B-Instruct-2507 on contexts of five patients (about 60k tokens) compacted in five chunks, and give Llama-3.1-8B results only as curves (their Figure~4).
Their released configuration for this setting also gives each head its own budget: a fixed per-head schedule, capped at 75\% of a head's keys, that averages to the target ratio, where we keep the same share of keys in every head.
With that schedule the AM-style fit reaches 67.8 and 65.2\%, which closes most of its own gap and about half of our fit's (Appendix~\ref{app:budget}).
The NNLS weighting does not: their fit matches each query's raw attention mass, ours its share relative to the row sum, and switching ours to theirs changes the AM-style fit's KL by $+0.17$ $[-0.35,+0.72]$ and $+0.19$ $[-0.42,+0.75]$ at 10\% and 5\% on the first 8 articles (146 questions).

\subsection{Per-head budgets}
\label{app:budget}
We reran Qwen3-4B on QuALITY with the published per-head schedule and the AM-style fit with their NNLS weighting, on a Tesla T4 in bf16, and four more settings with each model's published schedule on an A100 in bf16 (Table~\ref{tab:app-budget}): LongHealth with Qwen3-4B, with Llama-3.1-8B-Instruct and with Qwen3-4B-Instruct-2507, the model of the published LongHealth setting, and QuALITY with Llama-3.1-8B-Instruct.
Keep and drop take the same per-head numbers of keys, and the fold folds the AM fit head by head.

In the three settings with an equal-budget reference every method gains from the schedule, far more on LongHealth than on QuALITY.
On Qwen3-4B QuALITY AM now loses 1.3 and 3.9 points to the full cache at 10\% and 5\%, against 7.5 and 15.2 for the AM-style fit with equal budgets per head.
On Qwen3-4B LongHealth, against the AM-style fit with equal budgets per head on the same GPU type, whose full-cache answers are identical, the schedule raises AM's accuracy by 27.6 and 31.7 points at 10\% and 5\%, so that AM loses 3.3 and 6.8 points to the full cache instead of 30.9 and 38.5 (Table~\ref{tab:app-lh-recipe}).
Runs on different GPU types are compared only through each method's own accuracy and KL.

The fold is level with AM in nine of the ten cells: no fold$-$AM difference in accuracy, agreement or KL is beyond noise.
The exception is Llama-3.1-8B on LongHealth at 5\%, where the fold loses 4.5 points to AM, $[-6.5,-2.5]$, agrees with the full cache 6.0 points less often and has a KL higher by 0.16, all beyond noise; Appendix~\ref{app:llama-lh} diagnoses this failure.
The signs are consistent, though: in all ten cells the fold's KL is above AM's and its agreement below.
Under these budgets the bias itself is hard to detect.
AM's KL is below keep's in every cell, never beyond noise; AM is above keep beyond noise only in agreement on Llama QuALITY at 10\%, and its accuracy gains on Qwen3-4B LongHealth, 3.0 and 3.5 points, exclude zero but not by our margin.
No drop$-$fold KL difference is beyond noise in any cell.
In accuracy drop is 3.5 points below the fold on Qwen3-4B LongHealth at 5\% and 7.8 points above it on Llama LongHealth at 5\%, both beyond noise, and in that Llama cell keep and drop both score above AM, within noise.

\begin{table}[!htbp]
\centering
\footnotesize
\caption{The fold under each model's published per-head budgets: level with AM in nine of the ten cells, and below it on Llama-3.1-8B LongHealth at 5\%. AM-style fit with the published NNLS weighting, bf16; Qwen3-4B QuALITY on a Tesla T4, the rest on an A100. Below each paired difference, its 95\% bootstrap interval over the 30 articles (534 questions) or 20 patients (398); accuracy and agreement in \%, differences in points. Bold: beyond noise.}
\label{tab:app-budget}
\setlength{\tabcolsep}{4pt}
\begin{tabular}{lrrrrrr}
\toprule
 & \multicolumn{3}{c}{10\%} & \multicolumn{3}{c}{5\%} \\
\cmidrule(lr){2-4}\cmidrule(lr){5-7}
 & acc & agree & KL & acc & agree & KL \\
\midrule
\multicolumn{7}{l}{\emph{Qwen3-4B, QuALITY; full cache 69.1\%}} \\
keep & 69.5 & 84.5 & 1.18 & 63.3 & 77.5 & 1.93 \\
drop & 68.7 & 86.0 & 1.14 & 64.4 & 79.4 & 1.65 \\
AM & 67.8 & 86.7 & 0.90 & 65.2 & 81.1 & 1.65 \\
fold & 68.5 & 86.5 & 0.96 & 64.8 & 80.9 & 1.80 \\
fold$-$AM & +0.7 & $-$0.2 & +0.06 & $-$0.4 & $-$0.2 & +0.15 \\
 & {\tiny [$-$1.1, +2.7]} & {\tiny [$-$1.8, +1.5]} & {\tiny [$-$0.12, +0.25]} & {\tiny [$-$2.1, +1.7]} & {\tiny [$-$1.9, +1.6]} & {\tiny [$-$0.03, +0.31]} \\
drop$-$fold & +0.2 & $-$0.6 & +0.19 & $-$0.4 & $-$1.5 & $-$0.16 \\
 & {\tiny [$-$2.4, +2.6]} & {\tiny [$-$3.6, +2.4]} & {\tiny [$-$0.08, +0.48]} & {\tiny [$-$3.1, +2.2]} & {\tiny [$-$4.9, +1.7]} & {\tiny [$-$0.41, +0.11]} \\
AM$-$keep & $-$1.7 & +2.2 & $-$0.28 & +1.9 & +3.6 & $-$0.28 \\
 & {\tiny [$-$4.1, +0.8]} & {\tiny [$-$0.6, +5.2]} & {\tiny [$-$0.63, +0.03]} & {\tiny [$-$1.5, +5.7]} & {\tiny [$-$0.2, +7.1]} & {\tiny [$-$0.55, $-$0.02]} \\
\midrule
\multicolumn{7}{l}{\emph{Qwen3-4B, LongHealth; full cache 77.4\%}} \\
keep & 71.1 & 84.7 & 1.28 & 67.1 & 79.4 & 1.96 \\
drop & 71.6 & 88.4 & 1.13 & 68.1 & 80.4 & 2.08 \\
AM   & 74.1 & 88.4 & 0.97 & 70.6 & 82.9 & 1.71 \\
fold & 73.1 & 87.9 & 1.30 & 71.6 & 81.9 & 1.79 \\
fold$-$AM & $-$1.0 & $-$0.5 & +0.32 & +1.0 & $-$1.0 & +0.08 \\
 & {\tiny [$-$2.8, +0.5]} & {\tiny [$-$2.5, +1.3]} & {\tiny [+0.02, +0.66]} & {\tiny [$-$0.3, +2.3]} & {\tiny [$-$2.8, +0.8]} & {\tiny [$-$0.16, +0.29]} \\
drop$-$fold & $-$1.5 & +0.5 & $-$0.17 & \textbf{$-$3.5} & $-$1.5 & +0.29 \\
 & {\tiny [$-$4.0, +0.8]} & {\tiny [$-$2.8, +3.5]} & {\tiny [$-$0.58, +0.26]} & {\tiny [$-$5.8, $-$1.3]} & {\tiny [$-$4.5, +1.8]} & {\tiny [$-$0.11, +0.68]} \\
AM$-$keep & +3.0 & +3.8 & $-$0.30 & +3.5 & +3.5 & $-$0.24 \\
 & {\tiny [+1.0, +5.0]} & {\tiny [+0.3, +7.3]} & {\tiny [$-$0.68, +0.05]} & {\tiny [+1.0, +6.1]} & {\tiny [+0.0, +7.0]} & {\tiny [$-$0.61, +0.16]} \\
\midrule
\multicolumn{7}{l}{\emph{Qwen3-4B-Instruct-2507, LongHealth; full cache 82.2\%}} \\
keep & 79.6 & 89.9 & 0.46 & 77.9 & 85.4 & 0.87 \\
drop & 79.9 & 90.7 & 0.46 & 76.9 & 85.7 & 0.93 \\
AM   & 80.7 & 93.0 & 0.36 & 78.4 & 87.7 & 0.74 \\
fold & 80.2 & 91.0 & 0.42 & 77.9 & 86.2 & 0.77 \\
fold$-$AM & $-$0.5 & $-$2.0 & +0.07 & $-$0.5 & $-$1.5 & +0.03 \\
 & {\tiny [$-$2.0, +0.8]} & {\tiny [$-$3.5, $-$0.8]} & {\tiny [$-$0.01, +0.16]} & {\tiny [$-$2.3, +1.5]} & {\tiny [$-$3.3, +0.0]} & {\tiny [$-$0.04, +0.12]} \\
drop$-$fold & $-$0.3 & $-$0.3 & +0.04 & $-$1.0 & $-$0.5 & +0.16 \\
 & {\tiny [$-$2.0, +1.5]} & {\tiny [$-$2.0, +1.8]} & {\tiny [$-$0.18, +0.23]} & {\tiny [$-$3.0, +1.3]} & {\tiny [$-$2.3, +1.5]} & {\tiny [$-$0.02, +0.34]} \\
AM$-$keep & +1.0 & +3.0 & $-$0.11 & +0.5 & +2.3 & $-$0.13 \\
 & {\tiny [$-$2.0, +3.8]} & {\tiny [+1.0, +5.0]} & {\tiny [$-$0.28, +0.10]} & {\tiny [$-$2.0, +3.3]} & {\tiny [$-$1.0, +5.5]} & {\tiny [$-$0.31, +0.06]} \\
\midrule
\multicolumn{7}{l}{\emph{Llama-3.1-8B-Instruct, LongHealth; full cache 80.2\%}} \\
keep & 73.4 & 86.4 & 0.21 & 70.4 & 81.4 & 0.38 \\
drop & 73.6 & 86.2 & 0.23 & 70.9 & 79.6 & 0.39 \\
AM   & 73.1 & 86.9 & 0.17 & 67.6 & 77.9 & 0.32 \\
fold & 71.9 & 85.4 & 0.21 & 63.1 & 71.9 & 0.49 \\
fold$-$AM & $-$1.3 & $-$1.5 & +0.04 & \textbf{$-$4.5} & \textbf{$-$6.0} & \textbf{+0.16} \\
 & {\tiny [$-$3.8, +1.3]} & {\tiny [$-$4.0, +0.8]} & {\tiny [$-$0.01, +0.10]} & {\tiny [$-$6.5, $-$2.5]} & {\tiny [$-$8.1, $-$4.3]} & {\tiny [+0.13, +0.19]} \\
drop$-$fold & +1.8 & +0.8 & +0.02 & \textbf{+7.8} & \textbf{+7.8} & $-$0.09 \\
 & {\tiny [$-$1.5, +5.0]} & {\tiny [$-$2.8, +4.5]} & {\tiny [$-$0.06, +0.10]} & {\tiny [+3.3, +12.3]} & {\tiny [+2.8, +12.8]} & {\tiny [$-$0.17, $-$0.01]} \\
AM$-$keep & $-$0.3 & +0.5 & $-$0.03 & $-$2.8 & $-$3.5 & $-$0.06 \\
 & {\tiny [$-$4.0, +3.3]} & {\tiny [$-$3.5, +4.3]} & {\tiny [$-$0.07, +0.00]} & {\tiny [$-$5.8, +0.3]} & {\tiny [$-$7.3, +0.8]} & {\tiny [$-$0.14, +0.00]} \\
\bottomrule
\end{tabular}
\end{table}

\begin{table}[!htbp]
\centering
\footnotesize
\addtocounter{table}{-1}\renewcommand{\theHtable}{\arabic{table}.cont}
\caption{(continued) The fold under the published per-head budgets on Llama-3.1-8B-Instruct, QuALITY; layout as above.}
\setlength{\tabcolsep}{4pt}
\begin{tabular}{lrrrrrr}
\toprule
 & \multicolumn{3}{c}{10\%} & \multicolumn{3}{c}{5\%} \\
\cmidrule(lr){2-4}\cmidrule(lr){5-7}
 & acc & agree & KL & acc & agree & KL \\
\midrule
\multicolumn{7}{l}{\emph{Llama-3.1-8B-Instruct, QuALITY; full cache 71.3\%}} \\
keep & 69.3 & 85.0 & 0.19 & 67.6 & 82.8 & 0.28 \\
drop & 68.9 & 86.7 & 0.16 & 66.9 & 81.8 & 0.28 \\
AM   & 69.3 & 89.0 & 0.12 & 66.7 & 84.6 & 0.22 \\
fold & 68.9 & 88.4 & 0.14 & 66.7 & 83.0 & 0.23 \\
fold$-$AM & $-$0.4 & $-$0.6 & +0.02 & +0.0 & $-$1.7 & +0.01 \\
 & {\tiny [$-$2.3, +1.7]} & {\tiny [$-$3.0, +1.8]} & {\tiny [+0.01, +0.03]} & {\tiny [$-$1.7, +1.9]} & {\tiny [$-$4.3, +0.7]} & {\tiny [$-$0.00, +0.04]} \\
drop$-$fold & +0.0 & $-$1.7 & +0.01 & +0.2 & $-$1.1 & +0.04 \\
 & {\tiny [$-$3.4, +3.4]} & {\tiny [$-$5.2, +1.7]} & {\tiny [$-$0.01, +0.04]} & {\tiny [$-$2.8, +3.3]} & {\tiny [$-$3.9, +1.9]} & {\tiny [+0.00, +0.09]} \\
AM$-$keep & +0.0 & \textbf{+3.9} & $-$0.06 & $-$0.9 & +1.9 & $-$0.06 \\
 & {\tiny [$-$2.9, +3.1]} & {\tiny [+1.7, +6.3]} & {\tiny [$-$0.10, $-$0.03]} & {\tiny [$-$3.9, +2.2]} & {\tiny [$-$0.9, +5.1]} & {\tiny [$-$0.11, $-$0.02]} \\
\bottomrule
\end{tabular}
\end{table}

\subsection{The Llama-3.1-8B LongHealth failure}
\label{app:llama-lh}
Under the published per-head budgets the fold loses 4.5 points to AM on Llama-3.1-8B LongHealth at 5\% (Table~\ref{tab:app-budget}).
To locate the loss we reran the first five patients (100 questions) at 5\% with the same fit and budgets, on an A100 in bf16, and logged $q\cdot u$ and the attention outputs of every key-value head on the question's own queries (Table~\ref{tab:app-llama-lh}).
On these patients the rerun gives the same answers as the 20-patient run, and the fold is 5 points below AM, an interval that excludes zero but not by our noise margin, and 0.18 above it in KL, beyond noise.

Rounding the folded keys to bf16 is not the cause.
Kept in fp32 they leave the fold 4 points below AM and move its KL by 0.006, and the rounding term of the fold's logit error averages 0.0016 against 0.039 for the whole error, which equals its linear part $\beta\,(q\cdot u-1)$.
The question queries drift from the fit queries.
On the fit queries $q\cdot u$ has mean 1 by construction and an rms deviation from 1 of 0.044 on the average head; on the question queries the deviation is 0.071 and larger on all 1,280 heads (5 patients, 256 heads each), and on each question's last token, whose output gives the answer, $q\cdot u$ averages 0.96.
The biases the fold must carry are large: 36\% of the kept keys sit at the box $|\beta|=3$, and under AM's attention on the question queries $\beta$ has a mean of 0.62 and a variance of 0.79.
But the drift alone does not explain the loss, which needs the value refit: on AM's original values the fold is level with AM on the same values, and it scores 7 points more than the refitted fold.

Two changes each remove the gap.
Fitting on self-study queries as well as the repeat prefill, as the published AM does, leaves the fold level with AM; it also changes the kept keys, which are chosen on the same queries.
Boxing $\beta$ to $\pm1$ does the same and raises AM from 62 to 71\% and the fold from 57 to 70\%.
These runs cover five patients, with intervals over five clusters, so they locate the failure but do not by themselves confirm either change.

We reran the $\pm1$ box on all 20 patients (398 questions), with the fit and budgets otherwise as in the 20-patient run, on a different A100 (Table~\ref{tab:app-llama-lh}, lower panel); on patients 0--4 it reproduces the five-patient $\pm1$ run exactly.
The gap disappears: the fold answers more questions correctly than AM on 7 patients, fewer on 3 and as many on 10, against 1, 13 and 6 with the $\pm3$ box.
Against the $\pm3$ run, AM gains 5.3 points, $[+2.5,+7.8]$, and the fold 10.8, $[+8.0,+13.6]$, both beyond noise; the fold's agreement with the full cache rises by 10.6 points and its KL falls by 0.17, while AM's KL does not move ($+0.002$, $[-0.030,+0.035]$).
The fold now scores 3.5 points above keep, $[+1.3,+5.8]$, beyond noise; AM is 2.5 above keep, $[+0.3,+4.8]$, which does not clear our margin.
The question queries drift as much as before: rms$|q\cdot u-1|$ is 0.072 on them against 0.045 on the fit queries, larger on all 5,120 heads.
What changes is $\beta$: under AM's attention on the question queries its variance is 0.086, against 0.79 with the $\pm3$ box on patients 0--4 (the 20-patient $\pm3$ run did not log these diagnostics), and on the median head the fold moves AM's output by 0.017 of its norm, against 0.066.
We did not rerun the self-study queries on all 20 patients.

\begin{table}[t]
\centering
\small
\caption{Diagnosing the fold's loss on Llama-3.1-8B LongHealth at 5\%. Upper panel: patients 0--4 (100 questions, so accuracies are whole points), full-cache accuracy 79\%. Lower panel: all 20 patients (398 questions), full-cache accuracy 80.2\%; the $|\beta|\le3$ row is the run of Table~\ref{tab:app-budget}, and the $|\beta|\le1$ row ran on another A100 with identical full-cache, keep and drop log-probs. Published per-head budgets, AM-style fit with the published NNLS weighting, A100, bf16. AM's values: AM and the fold both keep AM's original values instead of refitting them. Fold$-$AM: paired difference with its 95\% bootstrap interval over the patients. Bold: beyond noise.}
\label{tab:app-llama-lh}
\begin{tabular}{lrrrrrr}
\toprule
 & \multicolumn{4}{c}{accuracy (\%)} & \multicolumn{2}{c}{fold$-$AM} \\
\cmidrule(lr){2-5}\cmidrule(lr){6-7}
run & keep & drop & AM & fold & acc & KL \\
\midrule
as in the main runs ($|\beta|\le3$) & 65 & 63 & 62 & 57 & $-$5 {\tiny [$-$9, $-$1]} & \textbf{+0.18} {\tiny [+0.14, +0.23]} \\
folded keys in fp32 & 65 & -- & 62 & 58 & $-$4 {\tiny [$-$7, $-$1]} & \textbf{+0.19} {\tiny [+0.14, +0.24]} \\
AM's values & 65 & -- & 61 & 64 & +3 {\tiny [$-$1, +5]} & $-$0.01 {\tiny [$-$0.05, +0.03]} \\
self-study fit queries & 69 & 65 & 64 & 64 & +0 {\tiny [$-$4, +4]} & +0.03 {\tiny [$-$0.03, +0.11]} \\
$|\beta|\le1$ & 65 & 63 & 71 & 70 & $-$1 {\tiny [$-$6, +5]} & $-$0.04 {\tiny [$-$0.09, +0.01]} \\
\midrule
\multicolumn{7}{l}{\emph{all 20 patients}} \\
$|\beta|\le3$ & 70.4 & 70.9 & 67.6 & 63.1 & \textbf{$-$4.5} {\tiny [$-$6.5, $-$2.5]} & \textbf{+0.16} {\tiny [+0.13, +0.19]} \\
$|\beta|\le1$ & 70.4 & 70.9 & 72.9 & 73.9 & +1.0 {\tiny [$-$1.0, +3.0]} & $-$0.01 {\tiny [$-$0.04, +0.02]} \\
\bottomrule
\end{tabular}
\end{table}

\subsection{The fold against dropping the bias}
\label{app:corollary}
Dropping $\beta$ while keeping AM's keys and values is the fold with $q\cdot u=0$, that is $t(q)=-1$ (Section~\ref{sec:method-bound}).
To compare the two, let $g(s)=o(\ell^{\mathrm{AM}}+s\,\beta)-o^{\mathrm{AM}}$ be the change in output when every bias of the head is scaled by $1+s$, so that $o^{\mathrm{fold}}-o^{\mathrm{AM}}=g(t(q))$ and $o^{\mathrm{drop}}-o^{\mathrm{AM}}=g(-1)$.

\smallskip\noindent\textbf{Corollary.}
Fix a query $q$ with $|t(q)|\le1$ and use the same values in all three caches.
Let $\kappa$ be the largest variance of $\beta$ under the attention weights $p(s)=\operatorname{softmax}(\ell^{\mathrm{AM}}+s\beta)$ for $s\in[-1,1]$, so $\kappa\le9$.
Then
\begin{equation}
\Big|\,\big\|o^{\mathrm{fold}}-o^{\mathrm{AM}}\big\|-|q\cdot u-1|\,\big\|o^{\mathrm{drop}}-o^{\mathrm{AM}}\big\|\,\Big|\;\le\;\tfrac12\,|t(q)|\,\big(1+|t(q)|\big)\,\kappa\,D_V ,
\label{eq:corollary}
\end{equation}
so to first order in $t(q)$ and in the spread of the attended biases, the fold's error is $|q\cdot u-1|$ times the error of dropping $\beta$.
The bounds agree with this: \eqref{eq:bound} gives $\tfrac32D_V$ for drop and $\tfrac32D_V|q\cdot u-1|$ for the fold.

\smallskip\noindent\emph{Proof sketch.}
By \eqref{eq:deriv}, $g(0)=0$ and $g'(0)=\operatorname{Cov}_{p(0)}(\beta,v)$.
Differentiating once more gives $g''(s)=\sum_j p_j(s)\,(\beta_j-\bar\beta_s)^2\,(v_j-\bar v_s)$, whose norm is at most $\operatorname{Var}_{p(s)}(\beta)\,D_V\le\kappa D_V$.
Taylor's theorem then gives $\|g(t)-t\,g'(0)\|\le\tfrac12t^2\kappa D_V$ and $\|g(-1)+g'(0)\|\le\tfrac12\kappa D_V$.
Combining the two with the triangle inequality gives \eqref{eq:corollary}.
\hfill$\square$

The corollary holds for fixed values, so it describes the fold before the refit of Section~\ref{sec:method-refit} and compares it with dropping $\beta$ from AM's cache; the \emph{drop} baseline of Section~\ref{sec:downstream} instead refits its values for $\beta=0$.
Its right side is small next to the drop error only when the attended biases vary little.
The diagnostic run of Appendix~\ref{app:llama-lh}, on Llama-3.1-8B LongHealth at 5\%, gives a weaker check: it logs head means over the question queries, each cache keeps its own refitted values, and neither $D_V$ nor $\kappa$ is logged, so the bound itself cannot be evaluated.
Over 1,280 heads per run, the ratio of the fold's error to drop's rises with rms$|q\cdot u-1|$, with Spearman correlations of 0.52 and 0.38 under the $\pm3$ box (two sets of fit queries) and 0.85 under a $\pm1$ box.
On the median head the fold's error is 0.13 of drop's, and 2.0 times the first-order prediction, rms$|q\cdot u-1|$ times drop's error, with $|\beta|\le3$ and 1.1 times with $|\beta|\le1$, where the attention-weighted variance of $\beta$, a lower bound on $\kappa$, is a tenth as large.

\subsection{Folding other biased caches}
\label{app:beyond}
The fold applies structurally to any static logit term (Table~\ref{tab:census}), and we test it on one outside AM: log-count merging, where a kept key absorbs evicted keys and its logit gains $\beta=\log n$ for a cluster of $n$ keys~\citep{bolya2022tome}.
We rerun Qwen3-4B with the published per-head budgets and the AM-style fit of Appendix~\ref{app:budget}, on QuALITY on a Tesla T4 and on LongHealth on an A100, in bf16 (Table~\ref{tab:merge}).
Every merge variant keeps the same keys as keep, which scores exactly as in Table~\ref{tab:app-budget}.
An evicted key joins the kept key whose post-RoPE key is most similar if their cosine is at least 0.9, and is dropped otherwise.
A cluster holds at most 16 keys, counting the kept one, so $\beta\le\log16=2.77$.
\emph{Merge-drop} gives each kept key its cluster's mean value and no bias, and \emph{merge-log} adds $\beta=\log n$.
\emph{Merge-fold} folds that $\beta$ into the key with $u$ from \eqref{eq:u} and keeps the mean values, and \emph{merge-refit} also refits the values by \eqref{eq:vrefit}.

On both benchmarks the fold reproduces merge-log: every merge-fold$-$merge-log interval contains zero.
Merge-log's KL is below merge-drop's in all four cells, beyond noise only on QuALITY at 5\%, and on LongHealth at 10\% its accuracy is 1.8 points lower, $[-3.3,-0.5]$, short of our margin.
Refitting the values raises merge-fold's accuracy in all four cells, within noise.

This is a test of the fold's fidelity to a second biased cache, since the gate merges few keys: 4.7\% and 2.5\% of the evicted keys on QuALITY at 10\% and 5\%, and 6.6\% and 3.7\% on LongHealth.
About three quarters of the kept keys on QuALITY absorb none and keep $\beta=0$.
With $\beta$ this small, the fold's logit error $\beta\,|q\cdot u-1|$ stays small: the attention-weighted mean logit gap between merge-fold and merge-log has a median over heads of 0.005 on QuALITY and 0.009 and 0.010 on LongHealth.
Without the cap, clusters grow to thousands of keys.
On the 37 questions over 2 articles that we used to choose the gate, the largest clusters at a cosine of 0.9 held 2,083 and 4,657 keys, a $\beta$ up to 8.4, and the fold's mean logit error on keys with $\beta>3$ had a median of 0.26 and 0.27 over heads.
We did not run that regime at scale.

We also fold two biased caches inside the code and harness of KV-Surgeon~\citep{wagner2026kvsurgeon}, changing nothing but the cache they return (Table~\ref{tab:e2}).
KV-Surgeon-fast selects keys by greedy elimination with closed-form compensation, then fits $\beta$ to the attention mass by box-constrained least squares, as AM does, and refits the values.
KVMerger~\citep{wang2024kvmerger} averages runs of adjacent similar keys and values with a Gaussian kernel and has no bias of its own; we use KV-Surgeon's variant of it, which fits $\beta$ to the attention mass in the same way.
Both keep $|\beta|\le3$.
The fold moves each kept key by $\sqrt{d}\,\beta u$, with $u$ from \eqref{eq:u} fitted on the compactor's own calibration queries, rounds the keys to bf16, sets $\beta=0$ and keeps the compactor's values.
In the caches the harness serves, no finite nonzero $\beta$ remains after the fold, and only the $-\infty$ entries that pad ragged heads are left in the mask.
We run Qwen3-4B on QuALITY at 5\% with KV-Surgeon's per-head budgets, on an A100 in bf16.
Unlike our other accuracy runs, the model generates its answer with thinking on, sampling with Qwen3's default settings for up to 2048 tokens, and KV-Surgeon's code parses the letter, counting unparsed answers as wrong.
Base and fold share a random seed per article, but their samples diverge once the logits differ, so part of each difference is sampling noise.
We ran 10 articles (181 questions per configuration).

Neither difference is beyond noise, and at 10 articles the intervals are 8 to 10 points wide.
The folded KVMerger cache leaves more answers unparsed, 11 against 5, all of them cut off at 2048 tokens, but this does not explain its deficit.
Four of the 16 questions it loses are unparsed under the fold, and four of the 11 it gains are unparsed under KVMerger; on the 166 questions parsed in both it gains 7 and loses 12.
The biases are larger than in the merge test: the median head reaches the box at $|\beta|=3$, and the mean $|\beta|$ per head averages 2.0 under KV-Surgeon-fast and 2.2 under KVMerger, while $|q\cdot u-1|$ has a median of 0.056 and 0.057 over heads.
We did not run the full cache or these compactors with $\beta$ dropped, so these runs measure only whether the fold works inside a third-party pipeline.

\begin{table}[!htbp]
\centering
\small
\caption{Log-count merging on Qwen3-4B: folding the count bias into the keys reproduces merge-log within noise on both benchmarks. Published per-head budgets, AM-style fit, cosine gate 0.9 on post-RoPE keys, at most 16 keys per cluster; bf16, QuALITY on a Tesla T4 and LongHealth on an A100. Differences: paired, with their 95\% bootstrap intervals over the 30 articles or 20 patients; accuracy and agreement in \%, differences in points. Bold: beyond noise. QuALITY: 534 questions, full-cache accuracy 69.1\%; LongHealth: 398 questions, 77.4\%. Keep, on the same keys, is in Table~\ref{tab:app-budget}.}
\label{tab:merge}
\begin{tabular}{lrrrrrr}
\toprule
 & \multicolumn{3}{c}{10\%} & \multicolumn{3}{c}{5\%} \\
\cmidrule(lr){2-4}\cmidrule(lr){5-7}
 & acc & agree & KL & acc & agree & KL \\
\midrule
\multicolumn{7}{l}{\emph{QuALITY}} \\
merge-drop  & 69.1 & 84.3 & 1.15 & 62.9 & 76.4 & 1.95 \\
merge-log   & 67.8 & 86.0 & 0.97 & 64.2 & 80.0 & 1.72 \\
merge-fold  & 67.2 & 85.4 & 0.98 & 64.4 & 80.9 & 1.74 \\
merge-refit & 68.5 & 86.3 & 0.90 & 66.3 & 81.1 & 1.50 \\
\midrule
merge-fold$-$merge-log & $-$0.6 & $-$0.6 & +0.01 & +0.2 & +0.9 & +0.01 \\
 & {\tiny [$-$1.5, +0.2]} & {\tiny [$-$1.5, +0.2]} & {\tiny [$-$0.02, +0.04]} & {\tiny [$-$1.1, +1.5]} & {\tiny [$-$0.6, +2.4]} & {\tiny [$-$0.04, +0.06]} \\
merge-log$-$merge-drop & $-$1.3 & +1.7 & $-$0.19 & +1.3 & +3.6 & \textbf{$-$0.23} \\
 & {\tiny [$-$3.8, +1.5]} & {\tiny [$-$0.8, +4.4]} & {\tiny [$-$0.36, $-$0.02]} & {\tiny [$-$1.1, +3.6]} & {\tiny [+0.6, +6.6]} & {\tiny [$-$0.41, $-$0.06]} \\
merge-refit$-$merge-fold & +1.3 & +0.9 & $-$0.08 & +1.9 & +0.2 & $-$0.24 \\
 & {\tiny [$-$0.8, +3.4]} & {\tiny [$-$0.8, +2.7]} & {\tiny [$-$0.21, +0.05]} & {\tiny [$-$0.4, +4.2]} & {\tiny [$-$2.9, +3.3]} & {\tiny [$-$0.48, $-$0.01]} \\
\midrule
\multicolumn{7}{l}{\emph{LongHealth}} \\
merge-drop  & 72.1 & 86.7 & 1.24 & 67.3 & 79.4 & 1.99 \\
merge-log   & 70.4 & 85.4 & 1.19 & 68.3 & 78.6 & 1.88 \\
merge-fold  & 70.6 & 85.9 & 1.16 & 67.8 & 78.6 & 1.88 \\
merge-refit & 72.9 & 89.2 & 1.04 & 69.6 & 81.7 & 1.88 \\
\midrule
merge-fold$-$merge-log & +0.3 & +0.5 & $-$0.02 & $-$0.5 & +0.0 & +0.00 \\
 & {\tiny [+0.0, +0.8]} & {\tiny [+0.0, +1.3]} & {\tiny [$-$0.09, +0.02]} & {\tiny [$-$2.0, +0.8]} & {\tiny [$-$1.5, +1.5]} & {\tiny [$-$0.05, +0.06]} \\
merge-log$-$merge-drop & $-$1.8 & $-$1.3 & $-$0.05 & +1.0 & $-$0.8 & $-$0.11 \\
 & {\tiny [$-$3.3, $-$0.5]} & {\tiny [$-$3.5, +1.0]} & {\tiny [$-$0.23, +0.15]} & {\tiny [$-$1.3, +3.3]} & {\tiny [$-$2.8, +1.3]} & {\tiny [$-$0.31, +0.13]} \\
merge-refit$-$merge-fold & +2.3 & +3.3 & $-$0.12 & +1.8 & +3.0 & $-$0.00 \\
 & {\tiny [$-$0.5, +5.3]} & {\tiny [+0.5, +6.3]} & {\tiny [$-$0.39, +0.16]} & {\tiny [$-$0.3, +4.0]} & {\tiny [+0.3, +5.8]} & {\tiny [$-$0.19, +0.21]} \\
\bottomrule
\end{tabular}
\end{table}

\begin{table}[!htbp]
\centering
\small
\caption{Folding the bias of two compactors inside KV-Surgeon's code and harness, Qwen3-4B QuALITY at 5\%: neither difference is beyond noise. KV-Surgeon's per-head budgets, answers generated with thinking on and parsed by its code, unparsed answers counted wrong; A100, bf16. Fold$-$base: paired difference in points with its 95\% bootstrap interval over the 10 articles. Gained/lost: questions right only with the fold / only without it. 181 questions.}
\label{tab:e2}
\begin{tabular}{lrrrrr}
\toprule
 & \multicolumn{2}{c}{accuracy (\%)} & & & unparsed \\
\cmidrule(lr){2-3}
base compactor & base & fold & fold$-$base & gained/lost & base/fold \\
\midrule
KV-Surgeon-fast & 68.5 & 71.8 & +3.3 {\tiny [$-$1.7, +8.2]} & 14/8 & 5/3 \\
KVMerger, fitted $\beta$ & 68.0 & 65.2 & $-$2.8 {\tiny [$-$6.8, +1.1]} & 11/16 & 5/11 \\
\bottomrule
\end{tabular}
\end{table}

\subsection{Ablations}
\label{app:abl}
We change one choice at a time on QuALITY, with Qwen3-1.7B, 4B and 8B at budgets of 10\% and 5\% and the protocol of Section~\ref{sec:downstream}.
Accuracy intervals are 3 to 7 points wide on 4B and 8B and up to 11 on 1.7B, so KL carries most of the comparison.

\begin{table}[!htbp]
\centering
\caption{On 4B and 8B the value refit and the covariance term in $u$ each lower the fold's KL. On 1.7B under our fit both raise it. Each row is a variant minus the cache it modifies, paired within one ablation run, with its 95\% bootstrap interval over the 30 articles below it. QuALITY, our AM fit. \emph{Mean dir.}: $u=\mu/\|\mu\|^2$ instead of \eqref{eq:u}. Absolute values for the default caches are in Tables~\ref{tab:downstream-quality} and~\ref{tab:app-sweep}.}
\label{tab:abl-u}
\footnotesize
\setlength{\tabcolsep}{2.5pt}
\begin{tabular}{lcccccc}
\toprule
 & \multicolumn{2}{c}{Qwen3-1.7B} & \multicolumn{2}{c}{Qwen3-4B} & \multicolumn{2}{c}{Qwen3-8B} \\
\cmidrule(lr){2-3}\cmidrule(lr){4-5}\cmidrule(lr){6-7}
 & 10\% & 5\% & 10\% & 5\% & 10\% & 5\% \\
\midrule
\multicolumn{7}{l}{\emph{$\mathrm{KL}(\text{full}\,\|\,\cdot)$}} \\
AM: no refit $-$ refit           & $-$0.26 & $-$0.02 & +0.37 & +0.35 & +0.09 & +0.01 \\
                                 & {\tiny [$-$0.68, +0.14]} & {\tiny [$-$0.42, +0.39]} & {\tiny [+0.12, +0.64]} & {\tiny [+0.09, +0.59]} & {\tiny [$-$0.11, +0.30]} & {\tiny [$-$0.27, +0.28]} \\
fold: no refit $-$ refit         & $-$0.53 & $-$0.21 & +0.43 & +0.44 & +0.36 & +0.67 \\
                                 & {\tiny [$-$0.97, $-$0.08]} & {\tiny [$-$0.67, +0.24]} & {\tiny [+0.09, +0.78]} & {\tiny [+0.13, +0.75]} & {\tiny [+0.08, +0.67]} & {\tiny [+0.33, +1.00]} \\
fold: mean dir.\ $-$ default     & $-$0.48 & $-$0.64 & +0.09 & +0.32 & +0.21 & +0.14 \\
                                 & {\tiny [$-$0.74, $-$0.21]} & {\tiny [$-$0.85, $-$0.44]} & {\tiny [$-$0.05, +0.24]} & {\tiny [+0.05, +0.58]} & {\tiny [+0.05, +0.37]} & {\tiny [+0.03, +0.25]} \\
no refit: fold $-$ AM            & +0.06 & +0.04 & +0.06 & +0.13 & +0.06 & +0.13 \\
                                 & {\tiny [$-$0.02, +0.15]} & {\tiny [$-$0.07, +0.14]} & {\tiny [+0.00, +0.13]} & {\tiny [+0.04, +0.22]} & {\tiny [+0.01, +0.12]} & {\tiny [+0.05, +0.21]} \\
\midrule
\multicolumn{7}{l}{\emph{accuracy (points)}} \\
AM: no refit $-$ refit           & +3.9 & +5.1 & $-$0.2 & +0.0 & $-$0.2 & +1.3 \\
                                 & {\tiny [+0.4, +7.8]} & {\tiny [+0.4, +10.2]} & {\tiny [$-$3.1, +2.7]} & {\tiny [$-$3.3, +3.2]} & {\tiny [$-$3.0, +2.7]} & {\tiny [$-$1.3, +3.9]} \\
fold: no refit $-$ refit         & +0.7 & +7.5 & $-$0.2 & +1.3 & $-$0.2 & +2.1 \\
                                 & {\tiny [$-$3.3, +5.4]} & {\tiny [+1.8, +12.8]} & {\tiny [$-$3.4, +3.4]} & {\tiny [$-$2.1, +4.8]} & {\tiny [$-$3.0, +2.6]} & {\tiny [+0.0, +4.3]} \\
fold: mean dir.\ $-$ default     & $-$1.3 & +5.2 & $-$0.4 & $-$0.4 & $-$0.4 & +0.6 \\
                                 & {\tiny [$-$3.7, +1.1]} & {\tiny [+1.7, +9.4]} & {\tiny [$-$2.1, +1.3]} & {\tiny [$-$2.6, +1.8]} & {\tiny [$-$2.4, +1.5]} & {\tiny [$-$2.1, +3.3]} \\
\bottomrule
\end{tabular}
\end{table}

\paragraph{Value refit.}
The refit has more to repair in the fold, whose logits on the reference queries miss AM's by $t(q_m)\beta_j$.
Without it the fold's KL on 4B and 8B sits slightly above AM's, and the refit lowers it by more than AM's, enough to put the fold below AM on 8B (Table~\ref{tab:abl-u}).
On 1.7B under our fit the refit hurts the fold instead (KL $+0.53$ at 10\%, 7.5 accuracy points at 5\%), and AM's numbers point the same way within noise.

\paragraph{Estimator for $u$.}
The mean direction also gives $q\cdot u=1$ on average over $Q_c$, but without the query covariance it cannot minimize the spread of $q\cdot u$.
On 4B and 8B it raises the fold's KL in all four columns, beyond noise in two, and on 8B at 10\% it cancels the fold's KL advantage over AM.
On 1.7B under our fit the order flips: the mean direction lowers KL and gains 5.2 accuracy points at 5\%.
Both 1.7B reversals occur where our fit is weak, with AM less accurate than keep at every budget of Table~\ref{tab:app-sweep}.
Under the AM-style fit, the stronger one on this model, the refit lowers KL for both caches (Appendix~\ref{app:fits}).
Because RoPE rotates queries by position while $u$ stays fixed, we also tried a $u$ restricted to the $d/4$ slowest-rotating pairs, which did not lower the fold's error in the layer study of Section~\ref{sec:why} and moved QuALITY accuracy by at most 1.1 points, so we fit $u$ over all coordinates.

\subsection{Other budgets and the AM-style fit}
\label{app:fits}
Table~\ref{tab:app-sweep} runs QuALITY with our fit at 20\% and 2\% on all four models, and also at 10\% and 5\% on Qwen3-1.7B.
Over the four budgets (10\% and 5\% in Table~\ref{tab:downstream-quality}) the fold's KL is below AM's on Qwen3-8B, more so as the budget shrinks, and every fold$-$AM KL interval on Qwen3-4B and Llama contains zero.
On Qwen3-1.7B our fit leaves AM below keep in accuracy at every budget, and the AM-style fit gives AM the lower KL (2.06 and 2.64 against 2.57 and 2.69 at 10\% and 5\%), which is why Table~\ref{tab:downstream-quality} uses it for this model.
In Table~\ref{tab:app-sweep} two drop$-$fold accuracy intervals exclude zero, neither by our margin: drop leads by 3.9 points on Qwen3-1.7B at 5\% ($[+0.6,+7.4]$) and trails by 4.1 on Qwen3-8B at 2\% ($[-7.4,-0.7]$).

\begin{table}[!htbp]
\centering
\caption{QuALITY under our AM fit, on all four models at 20\% and 2\% and on Qwen3-1.7B also at 10\% and 5\%: the fold's KL is below AM's beyond noise only on Qwen3-8B at 2\%, above it on Qwen3-1.7B at 20\% and 10\%, and level elsewhere. $\Delta$: fold minus AM, with the 95\% bootstrap interval over the 30 articles below it; accuracy in \%, $\Delta$acc in points. 534 questions; full-cache accuracy 55.6, 69.1, 73.0 and 71.3\% (55.6 here and 55.8 in Table~\ref{tab:downstream-quality}: run-to-run variation, Appendix~\ref{app:rerun}).}
\label{tab:app-sweep}
\footnotesize
\setlength{\tabcolsep}{2.5pt}
\begin{tabular}{llcccccccccc}
\toprule
 & & \multicolumn{4}{c}{accuracy (\%)} & \multicolumn{4}{c}{$\mathrm{KL}(\text{full}\,\|\,\cdot)$} & & \\
\cmidrule(lr){3-6}\cmidrule(lr){7-10}
model & budget & keep & drop & AM & fold & keep & drop & AM & fold & $\Delta$acc & $\Delta$KL \\
\midrule
Qwen3-1.7B & 20\% & 50.4 & 45.7 & 48.1 & 46.6 & 1.74 & 3.00 & 2.07 & 2.41 & $-$1.5 & +0.34 \\
 & &  &  &  &  &  &  &  &  & {\tiny [$-$4.4, +1.1]} & {\tiny [+0.13, +0.54]} \\
 & 10\% & 48.3 & 45.3 & 43.1 & 47.2 & 2.82 & 3.62 & 2.57 & 2.88 & +4.1 & +0.31 \\
 & &  &  &  &  &  &  &  &  & {\tiny [+1.7, +6.8]} & {\tiny [+0.08, +0.53]} \\
 & 5\% & 45.7 & 42.3 & 40.8 & 38.4 & 3.59 & 4.47 & 2.69 & 2.89 & $-$2.4 & +0.20 \\
 & &  &  &  &  &  &  &  &  & {\tiny [$-$5.4, +0.5]} & {\tiny [$-$0.06, +0.44]} \\
 & 2\% & 42.3 & 38.0 & 37.6 & 37.3 & 4.85 & 4.66 & 2.79 & 2.98 & $-$0.4 & +0.19 \\
 & &  &  &  &  &  &  &  &  & {\tiny [$-$3.5, +3.0]} & {\tiny [$-$0.06, +0.44]} \\
\midrule
Qwen3-4B & 20\% & 65.5 & 66.3 & 66.5 & 66.7 & 1.27 & 1.40 & 1.20 & 1.26 & +0.2 & +0.06 \\
 & &  &  &  &  &  &  &  &  & {\tiny [$-$1.1, +1.5]} & {\tiny [$-$0.05, +0.17]} \\
 & 2\% & 49.4 & 48.9 & 53.4 & 51.3 & 3.72 & 4.50 & 3.73 & 3.73 & $-$2.1 & $-$0.00 \\
 & &  &  &  &  &  &  &  &  & {\tiny [$-$4.9, +0.7]} & {\tiny [$-$0.22, +0.20]} \\
\midrule
Qwen3-8B & 20\% & 71.5 & 69.3 & 70.4 & 71.0 & 0.91 & 0.97 & 0.70 & 0.64 & +0.6 & $-$0.06 \\
 & &  &  &  &  &  &  &  &  & {\tiny [$-$0.9, +2.1]} & {\tiny [$-$0.14, +0.01]} \\
 & 2\% & 54.5 & 53.2 & 56.9 & 57.3 & 3.10 & 3.09 & 2.56 & 2.03 & +0.4 & $-$0.53 \\
 & &  &  &  &  &  &  &  &  & {\tiny [$-$2.1, +3.0]} & {\tiny [$-$0.75, $-$0.32]} \\
\midrule
Llama-3.1-8B & 20\% & 68.2 & 68.9 & 68.7 & 68.9 & 0.22 & 0.25 & 0.21 & 0.21 & +0.2 & +0.00 \\
 & &  &  &  &  &  &  &  &  & {\tiny [$-$1.1, +1.5]} & {\tiny [$-$0.01, +0.02]} \\
 & 2\% & 57.3 & 54.9 & 56.7 & 55.4 & 0.74 & 0.86 & 0.71 & 0.72 & $-$1.3 & +0.01 \\
 & &  &  &  &  &  &  &  &  & {\tiny [$-$3.1, +0.6]} & {\tiny [$-$0.02, +0.04]} \\
\bottomrule
\end{tabular}
\end{table}

Table~\ref{tab:abl-recipe} gives the AM-style fit on QuALITY for Qwen3-4B and 8B, and with self-study queries for 1.7B; Table~\ref{tab:app-lh-recipe} gives it on LongHealth.
On QuALITY it leaves AM above keep in KL on Qwen3-4B and shrinks the fold's KL advantage on 8B to the edge of our noise rule.
On Qwen3-1.7B, where it is the stronger fit, adding self-study queries narrows the fold's KL gap to AM from 0.32 and 0.33 to 0.14 and 0.21, and the fold then keeps 91--92\% of AM's KL reduction over keep.
Under the AM-style fit without self-study queries, the value refit lowers KL for both caches on 1.7B: from 3.28 and 3.90 to 2.06 and 2.64 for AM at 10\% and 5\%, and from 3.15 and 3.79 to 2.38 and 2.97 for the fold (Table~\ref{tab:downstream-quality}).
Across the fourteen Qwen3 QuALITY runs at 10\% and 5\% in Tables~\ref{tab:downstream-quality}, \ref{tab:app-sweep} and~\ref{tab:abl-recipe}, only one fold$-$AM accuracy interval excludes zero, and it favors the fold: $+4.1$ points $[+1.7,+6.8]$ on 1.7B under our fit at 10\%.
That interval clears our margin, so the cell is beyond noise in the fold's favor; a repeat run gives $+3.0$ $[+0.6,+5.5]$, short of the margin (Appendix~\ref{app:rerun}).

\begin{table}[!htbp]
\centering
\caption{QuALITY under the AM-style fit, on Qwen3-4B and 8B and with self-study queries on 1.7B: the fold's KL is lower than AM's on 8B, at the edge of noise, level on 4B, and higher on 1.7B with self-study queries, beyond noise at 5\%. SS (self-study) adds the queries of AM's four self-study prompts to $Q$. Keep differs from Table~\ref{tab:downstream-quality} because the kept keys depend on the fit's ranking and, with SS, on $Q$. $\Delta$: fold minus AM with 95\% paired bootstrap interval over the 30 articles.}
\label{tab:abl-recipe}
\footnotesize
\setlength{\tabcolsep}{3pt}
\begin{tabular}{lllccccc}
\toprule
 & & & \multicolumn{3}{c}{$\mathrm{KL}(\text{full}\,\|\,\cdot)$} & & \\
\cmidrule(lr){4-6}
Qwen3 & fit & budget & keep & AM & fold & $\Delta$KL & $\Delta$acc (points) \\
\midrule
1.7B & AM-style + SS & 10\% & 3.78 & 2.11 & 2.26 & +0.14 [$-$0.00, +0.30] & +0.9 [$-$0.7, +2.4] \\
 &  & 5\% & 4.97 & 2.44 & 2.65 & +0.21 [+0.07, +0.37] & $-$0.9 [$-$3.4, +1.7] \\
4B & AM-style & 10\% & 2.10 & 2.40 & 2.38 & $-$0.02 [$-$0.20, +0.17] & $-$0.6 [$-$2.5, +1.5] \\
 &  & 5\% & 2.88 & 3.20 & 3.21 & +0.01 [$-$0.16, +0.16] & +1.1 [$-$0.8, +3.0] \\
8B & AM-style & 10\% & 1.48 & 1.35 & 1.18 & $-$0.16 [$-$0.30, $-$0.03] & +0.4 [$-$1.5, +2.2] \\
 &  & 5\% & 2.38 & 1.98 & 1.83 & $-$0.15 [$-$0.28, $-$0.02] & $-$0.2 [$-$1.9, +1.5] \\
\bottomrule
\end{tabular}
\end{table}

On LongHealth the AM-style fit serves AM worse than ours: it gives AM a higher KL at every size and budget, and on Qwen3-4B a higher KL than keep.
Keep's accuracy rises in all six cells, leaving AM at or below it in accuracy everywhere.
Only one fold$-$AM accuracy interval excludes zero, on Qwen3-1.7B at 10\%, where the fold is 3.0 points lower, $[-5.5,-0.5]$, short of our margin; under our fit the same cell is a loss beyond noise.
The fold$-$AM KL interval excludes zero only on Qwen3-8B at 5\%, where the fold is lower.

\begin{table}[!htbp]
\centering
\caption{LongHealth under the AM-style fit (our fit: Table~\ref{tab:downstream-longhealth}): AM's KL is higher than under our fit at every size and budget, and its accuracy is at or below keep's in every cell. Columns as in Table~\ref{tab:app-sweep}; intervals over the 20 patients, 398 questions. Full-cache accuracy 68.3, 77.4 and 80.4\%.}
\label{tab:app-lh-recipe}
\footnotesize
\setlength{\tabcolsep}{2.2pt}
\begin{tabular}{llcccccccccc}
\toprule
 & & \multicolumn{4}{c}{accuracy (\%)} & \multicolumn{4}{c}{$\mathrm{KL}(\text{full}\,\|\,\cdot)$} & & \\
\cmidrule(lr){3-6}\cmidrule(lr){7-10}
model & budget & keep & drop & AM & fold & keep & drop & AM & fold & $\Delta$acc & $\Delta$KL \\
\midrule
Qwen3-1.7B & 10\% & 44.5 & 39.7 & 40.7 & 37.7 & 3.62 & 3.78 & 3.23 & 3.36 & $-$3.0 & +0.13 \\
 & &  &  &  &  &  &  &  &  & {\tiny [$-$5.5, $-$0.5]} & {\tiny [$-$0.07, +0.35]} \\
 & 5\% & 35.9 & 35.2 & 32.4 & 34.4 & 4.95 & 4.72 & 4.04 & 4.25 & +2.0 & +0.21 \\
 & &  &  &  &  &  &  &  &  & {\tiny [$-$1.0, +5.5]} & {\tiny [$-$0.02, +0.46]} \\
\midrule
Qwen3-4B & 10\% & 48.5 & 45.7 & 46.5 & 47.5 & 4.25 & 4.96 & 4.79 & 4.88 & +1.0 & +0.09 \\
 & &  &  &  &  &  &  &  &  & {\tiny [$-$2.0, +3.5]} & {\tiny [$-$0.18, +0.36]} \\
 & 5\% & 42.7 & 37.2 & 38.9 & 37.7 & 4.77 & 5.70 & 5.69 & 5.93 & $-$1.3 & +0.24 \\
 & &  &  &  &  &  &  &  &  & {\tiny [$-$3.8, +1.3]} & {\tiny [$-$0.03, +0.53]} \\
\midrule
Qwen3-8B & 10\% & 50.0 & 47.7 & 50.0 & 50.3 & 2.64 & 2.41 & 2.20 & 2.16 & +0.3 & $-$0.04 \\
 & &  &  &  &  &  &  &  &  & {\tiny [$-$2.3, +2.5]} & {\tiny [$-$0.14, +0.07]} \\
 & 5\% & 41.0 & 38.2 & 36.9 & 37.9 & 3.59 & 3.21 & 2.96 & 2.80 & +1.0 & $-$0.16 \\
 & &  &  &  &  &  &  &  &  & {\tiny [$-$1.5, +3.5]} & {\tiny [$-$0.26, $-$0.05]} \\
\bottomrule
\end{tabular}
\end{table}
\subsection{Agreement with the full cache}
Table~\ref{tab:app-agree} gives, for the main settings, the paired differences in the share of questions on which a cache's top option is the full cache's.
No fold$-$AM agreement difference is beyond noise except on LongHealth with Qwen3-1.7B at 10\%, where the fold is lower, and Qwen3-4B at 5\%, where it is higher.
Drop's agreement is below the fold's in 11 of the 12 settings where drop ran.

\begin{table}[!htbp]
\centering
\caption{Agreement with the full cache's top option: paired differences in points, with 95\% bootstrap intervals over contexts. AM fit as in Tables~\ref{tab:downstream-quality} and~\ref{tab:downstream-longhealth}; drop was not run with the AM-style fit on QuALITY.}
\label{tab:app-agree}
\footnotesize
\setlength{\tabcolsep}{3pt}
\begin{tabular}{llcccc}
\toprule
 & & \multicolumn{2}{c}{fold$-$AM} & \multicolumn{2}{c}{drop$-$fold} \\
\cmidrule(lr){3-4}\cmidrule(lr){5-6}
benchmark & model & 10\% & 5\% & 10\% & 5\% \\
\midrule
QuALITY & Qwen3-1.7B & +0.7 {\tiny [$-$1.7, +3.3]} & +0.6 {\tiny [$-$1.7, +3.1]} & -- & -- \\
 & Qwen3-4B & $-$1.3 {\tiny [$-$3.3, +0.7]} & $-$1.1 {\tiny [$-$3.4, +1.1]} & $-$3.6 {\tiny [$-$6.0, $-$1.1]} & $-$3.0 {\tiny [$-$6.2, +0.0]} \\
 & Qwen3-8B & $-$0.7 {\tiny [$-$3.2, +1.7]} & $-$0.6 {\tiny [$-$2.5, +1.3]} & $-$0.6 {\tiny [$-$2.4, +1.1]} & $-$1.3 {\tiny [$-$4.9, +2.1]} \\
 & Llama-3.1-8B & +0.6 {\tiny [$-$1.3, +2.6]} & +0.0 {\tiny [$-$1.5, +1.5]} & $-$0.4 {\tiny [$-$3.0, +2.4]} & $-$0.7 {\tiny [$-$4.9, +3.0]} \\
LongHealth & Qwen3-1.7B & $-$7.0 {\tiny [$-$12.2, $-$2.0]} & $-$1.5 {\tiny [$-$4.8, +1.8]} & +3.3 {\tiny [$-$0.5, +6.8]} & $-$1.0 {\tiny [$-$5.1, +3.3]} \\
 & Qwen3-4B & +2.0 {\tiny [$-$0.3, +4.0]} & +4.0 {\tiny [+2.3, +5.8]} & $-$1.0 {\tiny [$-$3.8, +1.8]} & $-$2.0 {\tiny [$-$5.3, +1.3]} \\
 & Qwen3-8B & +0.0 {\tiny [$-$3.5, +3.3]} & $-$1.0 {\tiny [$-$3.8, +1.8]} & $-$4.8 {\tiny [$-$8.8, $-$0.5]} & $-$1.3 {\tiny [$-$4.3, +2.0]} \\
\bottomrule
\end{tabular}
\end{table}

\subsection{Run-to-run variation}
\label{app:rerun}
The fitted caches do not reproduce bit for bit on the GPU, by enough to matter near zero.
We ran QuALITY with our fit at 10\% and 5\% twice on each Qwen3 model: once for Tables~\ref{tab:downstream-quality} and~\ref{tab:app-sweep} and once for the ablations of Table~\ref{tab:abl-u}.
Keep, which involves no fitting, gave the same answers in both, while AM and the fold differed by up to 4 of 534 questions in accuracy and by up to 0.03 in KL.
The fold--AM accuracy difference moved by up to 6 of 534 questions (1.1 points) and its agreement difference by up to 1.3.
That is enough to carry the 1.7B comparison at 5\% across zero, from $-2.4$ $[-5.4,+0.5]$ in Table~\ref{tab:app-sweep} to $-3.6$ $[-6.8,-0.6]$ in the repeat, a move of 6 questions between rounded entries; at 10\% the repeat gives $+3.0$ $[+0.6,+5.5]$ against $+4.1$.
Even the full cache moves by a question: Qwen3-1.7B scores 55.6\% in two runs and 55.8\% in a third, the AM-style run whose full cache Table~\ref{tab:downstream-quality} prints.
The bootstrap intervals cover sampling over contexts and miss this nondeterminism, which is where the noise rule of Section~\ref{sec:downstream} comes from.
Every other table reports the first of these runs; the repeat appears only here and in the paired differences of Table~\ref{tab:abl-u}.

\subsection{Serving details}

\paragraph{Stock attention kernels.}
\label{app:stock}
On stock bf16 kernels few answers change, and every fold--AM and drop--fold difference in accuracy or KL that was beyond noise stays beyond noise with the same sign.
The main text computes the question's attention over the cache with an explicit fp32 softmax; to check the fold's cache on a serving kernel, we reran QuALITY at 10\% under our fit on all four models through \texttt{scaled\_dot\_product\_attention} in bf16.
The full cache, keep, drop and the fold use the flash backend, and AM the memory-efficient or cuDNN backend with an additive mask.
Flash aligns its causal mask to the top left, so we prepend $r$ zero queries and discard their outputs; on random inputs this matches the fp32 path to within $9\times10^{-3}$.

Each variant picks the same option in both runs on 95.9 to 99.6\% of questions (Table~\ref{tab:app-stock}), and the fold's 0.6 to 2.1\% of changed answers is at most twice the full cache's 0.4 to 1.1\%.
The differences that flip sign were within $0.01$ KL of zero in both runs.
Llama-3.1-8B's fold--AM accuracy gain, whose interval touched zero, now has its lower end at $-1.1$ points.
Kernel numerics alone can move a question's KL even when its top option stays the same, so we compare KL only between caches run on the same kernel path.

\begin{table}[!htbp]
\centering
\caption{QuALITY at 10\%, our fit: the fp32 explicit-softmax path of the main text against stock PyTorch kernels in bf16 (flash without a bias, masked SDPA for AM). Accuracy and KL are fp32\,/\,stock. Same: share of questions with the same top option in both runs. Last two columns: fold$-$AM KL and drop$-$fold KL in the stock run, with 95\% bootstrap intervals.}
\label{tab:app-stock}
\small
\setlength{\tabcolsep}{3pt}
\begin{tabular}{lcccccc}
\toprule
 & \multicolumn{2}{c}{fold} & \multicolumn{2}{c}{AM} & \multicolumn{2}{c}{stock run} \\
\cmidrule(lr){2-3}\cmidrule(lr){4-5}\cmidrule(lr){6-7}
model & acc (\%) & same (\%) & acc (\%) & same (\%) & fold$-$AM KL & drop$-$fold KL \\
\midrule
Qwen3-1.7B & 47.2\,/\,46.4 & 97.9 & 43.1\,/\,43.6 & 95.9 & $+0.28$ [$+0.06$, $+0.49$] & $+0.71$ [$+0.42$, $+1.03$] \\
Qwen3-4B   & 62.7\,/\,63.3 & 98.7 & 63.7\,/\,63.9 & 99.6 & $-0.01$ [$-0.17$, $+0.15$] & $+0.68$ [$+0.44$, $+0.96$] \\
Qwen3-8B   & 69.1\,/\,69.3 & 98.9 & 68.4\,/\,68.9 & 99.1 & $-0.21$ [$-0.39$, $-0.05$] & $+0.59$ [$+0.40$, $+0.81$] \\
Llama-3.1-8B & 67.6\,/\,67.2 & 99.4 & 66.1\,/\,66.5 & 99.1 & $-0.00$ [$-0.02$, $+0.01$] & $+0.08$ [$+0.04$, $+0.12$] \\
\bottomrule
\end{tabular}
\end{table}

\paragraph{vLLM injection checks.}
\label{app:inject}
The connector writes the folded keys and values in place of $R$ placeholder tokens, for a compacted cache of $R$ entries from a context of $C$ tokens, through the backend's own cache-write operation.
The stored keys are rotated back by $C-R$ positions, so a question token at position $R+i$ sees the logits it would see at $C+i$ on the original keys.
For Qwen3-4B on QuALITY, reading the blocks back after the connector writes them gives a largest absolute difference of 0 over 232 injections.
The largest difference in an answer letter's log-probability between vLLM and Hugging Face is 0.33 for the full prompt and 0.22 and 0.28 for the folds at 10\% and 5\%.
For Qwen3-14B on LongBench, two uncompacted caches injected the same way reproduce Hugging Face's full-cache outputs exactly.
As controls on three samples, replacing the folded cache with zeros or shuffling its keys over tokens changes the output in five of six requests; the sixth, a zero cache, answers ``unanswerable'', as the fold does.

\paragraph{Kernel timing.}
\label{app:eager}
\begin{table}[!htbp]
\centering
\caption{With launch cost removed by CUDA graphs, AM's bias costs little in single-sequence decode and most at batch-8 decode. Top: 512--4096 cached entries, which covers our compacted caches. Bottom: 16k--32k. CUDA-graphed per-call times, one run per model. Cells: time over the fastest plain time (flash, cuDNN or FlexAttention) of the same model and setting, median over the four models and the $L$ values of the panel, min--max in parentheses. Biased: fastest implementation that accepts the bias. Flex: FlexAttention without a bias. Flex$+\beta$/Flex: FlexAttention with the bias over FlexAttention without it. Flash: per-call time of plain flash SDPA in $\mu$s, min--max.}
\label{tab:deploy}
\small
\setlength{\tabcolsep}{4pt}
\begin{tabular}{lcccc}
\toprule
 & \multicolumn{2}{c}{decode ($q_{\text{len}}=1$)} & \multicolumn{2}{c}{prefill ($q_{\text{len}}=512$)} \\
\cmidrule(lr){2-3}\cmidrule(lr){4-5}
 & $B=1$ & $B=8$ & $B=1$ & $B=8$ \\
\midrule
\multicolumn{5}{l}{\emph{$L=512$, 1024, 2048, 4096 entries}} \\
biased & 1.06 {\scriptsize(1.02--1.29)} & 1.58 {\scriptsize(1.09--1.90)} & 1.27 {\scriptsize(1.16--1.50)} & 1.26 {\scriptsize(1.16--1.39)} \\
Flex & 1.00 {\scriptsize(1.00--1.23)} & 1.29 {\scriptsize(1.00--1.45)} & 1.16 {\scriptsize(1.00--1.43)} & 1.16 {\scriptsize(1.07--1.24)} \\
Flex$+\beta$\,/\,Flex & 1.05 {\scriptsize(1.00--1.29)} & 1.19 {\scriptsize(1.09--1.33)} & 1.08 {\scriptsize(1.06--1.31)} & 1.08 {\scriptsize(1.06--1.30)} \\
flash ($\mu$s) & 13--22 & 19--92 & 52--344 & 130--1503 \\
\midrule
\multicolumn{5}{l}{\emph{$L=16$k and $32$k entries}} \\
biased & 1.15 {\scriptsize(1.07--1.26)} & 1.74 {\scriptsize(1.69--1.75)} & 1.44 {\scriptsize(1.43--1.70)} & 1.22 {\scriptsize(1.21--1.23)} \\
Flex & 1.00 {\scriptsize(1.00--1.01)} & 1.54 {\scriptsize(1.53--1.56)} & 1.37 {\scriptsize(1.35--1.57)} & 1.12 {\scriptsize(1.11--1.14)} \\
Flex$+\beta$\,/\,Flex & 1.14 {\scriptsize(1.07--1.26)} & 1.13 {\scriptsize(1.10--1.14)} & 1.07 {\scriptsize(1.07--1.11)} & 1.09 {\scriptsize(1.07--1.09)} \\
flash ($\mu$s) & 57--98 & 323--632 & 410--2182 & 3011--11926 \\
\bottomrule
\end{tabular}
\end{table}

We time one attention layer at each model's head shapes, all with $d=128$, on random bf16 tensors on an A100-SXM4-80GB with PyTorch~2.8, at 512 to 4096 cached entries per head and at 16k and 32k.
Each call attends 1 new query token (decode) or 512 (a prefill chunk) at batch size $B\in\{1,8\}$, and all biased paths use the same $\beta$, rounded once to bf16.
We capture 10 calls of each implementation in one CUDA graph and report the median per-call time over 50 replays.
Each model has one run, and Qwen3-1.7B lacks the 32k, $B=8$ prefill setting.
Of PyTorch's four SDPA backends, flash rejects every float mask, cuDNN runs neither cache at decode, and the memory-efficient backend takes grouped-query attention only after $K$ and $V$ are expanded to one copy per query head.
That expansion allocates up to 4,128\,MB per layer at 32k entries and $B=8$, about four times that layer's cache, and makes its biased decode 10 to 27 times slower than the fastest plain kernel at 16k--32k entries.
FlexAttention with the bias and cuDNN with the mask need no more memory above the cache than plain flash.
The largest error against an fp32 reference is $7.9\times10^{-3}$ for the biased paths and $2.0\times10^{-3}$ for the plain ones.
Flash is not always the fastest plain kernel at these sizes, and with up to 4096 entries the fastest biased path beats plain flash in 4 of the 16 single-sequence decode settings; of the 95 settings of Table~\ref{tab:deploy}, it is fastest in 52, cuDNN in 23 (all prefill) and FlexAttention in 20, so each cell divides by the fastest plain time.
In the same run, single eager calls at $B=1$ decode with up to 4096 entries took 84 to 88\,$\mu$s for flash and 172 to 185\,$\mu$s for FlexAttention, against 13 to 22 and 13 to 19\,$\mu$s under CUDA graphs.
Launch cost dominates these small calls without graphs, so we report only the graphed times, which leave out the projections, the MLP, the scheduler, paged-KV kernels and the one-time cost of the fold.

\subsection{Far-query diagnosis}
\label{app:far}
A $u$ fitted on $Q$ pins the mean of $q\cdot u$ to $1$ on the last quarter of the context only.
Fitting it on $Q_c$ removes 73, 53 and 66\% of the base fold's mean excess near the context on Qwen3-1.7B, 4B and 8B, and 54, 34 and 0\% far from it.
It works through the mean: the median $|m-1|$ near the context drops in every model, while the median CV of the two fits agrees to within $0.01$ at each distance.
On the worst far pairs it shrinks the drift without removing it, because no $u$ fitted on the context pins the mean thousands of tokens later.

The worst tenth of far pairs, ranked by the default fold's own excess, carries 90 to 96\% of that excess (Table~\ref{tab:why}), and 81 to 89\% of these pairs lie in layers 0 to 6.
On them the median $|m-1|$ is 0.18, 0.30 and 0.34, against 0.09, 0.12 and 0.11 on the other pairs, while the median $\sigma$ differs between the two groups by at most a factor of 1.5.

The mean over heads is worse than the median: on 4B and 8B the default fold's mean error is 1.1 to 2.6 times AM's from $\Delta=4$k on.
A few heads drive this: at $\Delta=16$k the fold's relative error reaches 98.6 on Qwen3-8B and 37.6 on 4B, both in layer 3, against a median of 0.47 and 0.41.
On 8B at $\Delta=16$k the ten worst heads, all in layers 0 to 6, carry three quarters of the summed fold error.

%% file: main.bbl
\begin{thebibliography}{33}
\providecommand{\natexlab}[1]{#1}
\providecommand{\url}[1]{\texttt{#1}}
\expandafter\ifx\csname urlstyle\endcsname\relax
  \providecommand{\doi}[1]{doi: #1}\else
  \providecommand{\doi}{doi: \begingroup \urlstyle{rm}\Url}\fi

\bibitem[Adams et~al.(2024)Adams, Busch, Han, Excoffier, Ortala, L{\"o}ser,
  Aerts, Kather, Truhn, and Bressem]{adams2024longhealth}
Lisa Adams, Felix Busch, Tianyu Han, Jean-Baptiste Excoffier, Matthieu Ortala,
  Alexander L{\"o}ser, Hugo J. W.~L. Aerts, Jakob~Nikolas Kather, Daniel Truhn,
  and Keno Bressem.
\newblock {LongHealth: A Question Answering Benchmark with Long Clinical
  Documents}, 2024.
\newblock URL \url{https://arxiv.org/abs/2401.14490}.

\bibitem[Anonymous(2026{\natexlab{a}})]{exactfinite}
Anonymous.
\newblock {Exact Finite-Query KV Cache Compilation}.
\newblock OpenReview, 2026{\natexlab{a}}.
\newblock URL \url{https://openreview.net/forum?id=OQhOBMNVeN}.
\newblock Submitted to ICLR 2027.

\bibitem[Anonymous(2026{\natexlab{b}})]{limits}
Anonymous.
\newblock {On the Limits of KV Cache Merging}.
\newblock OpenReview, 2026{\natexlab{b}}.
\newblock URL \url{https://openreview.net/forum?id=0YVzutZXGO}.
\newblock Submitted to ICLR 2027.

\bibitem[Anonymous(2026{\natexlab{c}})]{qream}
Anonymous.
\newblock {QREAM: KV Cache Compression via Pivoted QR and Explicit Attention
  Matching}.
\newblock OpenReview, 2026{\natexlab{c}}.
\newblock URL \url{https://openreview.net/forum?id=QJoILTys3B}.
\newblock Submitted to ICLR 2027.

\bibitem[Bai et~al.(2024)Bai, Lv, Zhang, Lyu, Tang, Huang, Du, Liu, Zeng, Hou,
  Dong, Tang, and Li]{bai2024longbench}
Yushi Bai, Xin Lv, Jiajie Zhang, Hongchang Lyu, Jiankai Tang, Zhidian Huang,
  Zhengxiao Du, Xiao Liu, Aohan Zeng, Lei Hou, Yuxiao Dong, Jie Tang, and
  Juanzi Li.
\newblock {LongBench: A Bilingual, Multitask Benchmark for Long Context
  Understanding}.
\newblock In \emph{Proceedings of the 62nd Annual Meeting of the Association
  for Computational Linguistics (Volume 1: Long Papers)}, pages 3119--3137,
  Bangkok, Thailand, 2024. Association for Computational Linguistics.
\newblock URL \url{https://aclanthology.org/2024.acl-long.172/}.

\bibitem[Bolya et~al.(2022)Bolya, Fu, Dai, Zhang, Feichtenhofer, and
  Hoffman]{bolya2022tome}
Daniel Bolya, Cheng-Yang Fu, Xiaoliang Dai, Peizhao Zhang, Christoph
  Feichtenhofer, and Judy Hoffman.
\newblock {Token Merging: Your ViT But Faster}, 2022.
\newblock URL \url{https://arxiv.org/abs/2210.09461}.

\bibitem[Bouyahiaoui et~al.(2026)Bouyahiaoui, laouar, Boutorh, and
  Ameur]{bouyahiaoui2026selkv}
Soumia Bouyahiaoui, Manel~Kara laouar, Aicha Boutorh, and Mohamed~Hadj Ameur.
\newblock {SelKV: Selective KV Cache Merging with Per-Token Merge-or-Drop and
  Attention Compensation}, 2026.
\newblock URL \url{https://arxiv.org/abs/2607.16213}.

\bibitem[Chen et~al.(2026)Chen, Shen, Xiao, and Han]{chen2026latent}
Yingfa Chen, Xingyu Shen, Chaojun Xiao, and Xu~Han.
\newblock {Latent Compaction with Test-Time Training}, 2026.
\newblock URL \url{https://arxiv.org/abs/2610.20582}.

\bibitem[Cui and Xu(2025)]{cui2025asymkv}
Wanyun Cui and Mingwei Xu.
\newblock {Homogeneous Keys, Heterogeneous Values: Exploiting Local KV Cache
  Asymmetry for Long-Context LLMs}, 2025.
\newblock URL \url{https://arxiv.org/abs/2506.05410}.

\bibitem[Dao(2023)]{dao2023flashattention2}
Tri Dao.
\newblock {FlashAttention-2: Faster Attention with Better Parallelism and Work
  Partitioning}, 2023.
\newblock URL \url{https://arxiv.org/abs/2307.08691}.

\bibitem[Kwon et~al.(2023)Kwon, Li, Zhuang, Sheng, Zheng, Yu, Gonzalez, Zhang,
  and Stoica]{kwon2023pagedattention}
Woosuk Kwon, Zhuohan Li, Siyuan Zhuang, Ying Sheng, Lianmin Zheng, Cody~Hao Yu,
  Joseph~E. Gonzalez, Hao Zhang, and Ion Stoica.
\newblock {Efficient Memory Management for Large Language Model Serving with
  PagedAttention}, 2023.
\newblock URL \url{https://arxiv.org/abs/2309.06180}.

\bibitem[Li et~al.(2026)Li, Li, and Lan]{li2026momentkv}
Yu~Li, Binxu Li, and Tian Lan.
\newblock {MomentKV: Closing the Directional Gap in KV Cache Eviction for
  Long-Context Inference}, 2026.
\newblock URL \url{https://arxiv.org/abs/2606.01563}.

\bibitem[Li et~al.(2024)Li, Huang, Yang, Venkitesh, Locatelli, Ye, Cai, Lewis,
  and Chen]{li2024snapkv}
Yuhong Li, Yingbing Huang, Bowen Yang, Bharat Venkitesh, Acyr Locatelli,
  Hanchen Ye, Tianle Cai, Patrick Lewis, and Deming Chen.
\newblock {SnapKV: LLM Knows What You are Looking for Before Generation}, 2024.
\newblock URL \url{https://arxiv.org/abs/2404.14469}.

\bibitem[Liu et~al.(2024{\natexlab{a}})Liu, Liu, Pan, He, Haffari, and
  Zhuang]{liu2024minicache}
Akide Liu, Jing Liu, Zizheng Pan, Yefei He, Gholamreza Haffari, and Bohan
  Zhuang.
\newblock {MiniCache: KV Cache Compression in Depth Dimension for Large
  Language Models}, 2024{\natexlab{a}}.
\newblock URL \url{https://arxiv.org/abs/2405.14366}.

\bibitem[Liu et~al.(2025)Liu, Wang, Liu, and Tang]{liu2025zsmerge}
Xin Liu, Xudong Wang, Pei Liu, and Guoming Tang.
\newblock {ZSMerge: Zero-Shot KV Cache Compression for Memory-Efficient
  Long-Context LLMs}, 2025.
\newblock URL \url{https://arxiv.org/abs/2503.10714}.

\bibitem[Liu et~al.(2026{\natexlab{a}})Liu, Chong, Zhang, Zheng, Liang, and
  Kefu]{liu2026tam}
Yang Liu, Bin Chong, Chongyang Zhang, Hao Zheng, Jiayu Liang, and Xu~Kefu.
\newblock {Thought-Aware KV Cache Compaction for Reasoning via Adaptive
  Attention Matching}, 2026{\natexlab{a}}.
\newblock URL \url{https://arxiv.org/abs/2608.12331}.

\bibitem[Liu et~al.(2026{\natexlab{b}})Liu, Ji, An, Jain, Polatkan, Zhu, and
  Chang]{liu2026online}
Yujian Liu, Jiabao Ji, Li~An, Rohit Jain, Gungor Polatkan, Siyu Zhu, and Shiyu
  Chang.
\newblock {Practical Online KV Cache Compaction for LLM Agents: An Empirical
  Study}, 2026{\natexlab{b}}.
\newblock URL \url{https://arxiv.org/abs/2608.00902}.

\bibitem[Liu et~al.(2024{\natexlab{b}})Liu, Yuan, Jin, Zhong, Xu, Braverman,
  Chen, and Hu]{liu2024kivi}
Zirui Liu, Jiayi Yuan, Hongye Jin, Shaochen Zhong, Zhaozhuo Xu, Vladimir
  Braverman, Beidi Chen, and Xia Hu.
\newblock {KIVI: A Tuning-Free Asymmetric 2bit Quantization for KV Cache},
  2024{\natexlab{b}}.
\newblock URL \url{https://arxiv.org/abs/2402.02750}.

\bibitem[O'Neill et~al.(2026)O'Neill, Sandomirsky, Partridge, Jayasekara, and
  Kirkby]{oneill2026still}
Charles O'Neill, Alex Sandomirsky, Harry Partridge, Mudith Jayasekara, and Max
  Kirkby.
\newblock {Still: Amortized KV Cache Compaction in a Single Forward Pass},
  2026.
\newblock URL \url{https://arxiv.org/abs/2606.07878}.

\bibitem[Pang et~al.(2022)Pang, Parrish, Joshi, Nangia, Phang, Chen,
  Padmakumar, Ma, Thompson, He, and Bowman]{pang2022quality}
Richard~Yuanzhe Pang, Alicia Parrish, Nitish Joshi, Nikita Nangia, Jason Phang,
  Angelica Chen, Vishakh Padmakumar, Johnny Ma, Jana Thompson, He~He, and
  Samuel~R. Bowman.
\newblock {QuALITY: Question Answering with Long Input Texts, Yes!}
\newblock In \emph{Proceedings of the 2022 Conference of the North American
  Chapter of the Association for Computational Linguistics: Human Language
  Technologies}, pages 5336--5358, Seattle, United States, 2022. Association
  for Computational Linguistics.
\newblock URL \url{https://aclanthology.org/2022.naacl-main.391/}.

\bibitem[Shen et~al.(2026)Shen, Wang, Tu, Zhang, Li, Aziz, Tang, and
  Li]{shen2026arckv}
Zheyu Shen, Guanhua Wang, Dezhan Tu, Mengchi Zhang, Yanjia Li, Adnan Aziz,
  Chunqiang Tang, and Ang Li.
\newblock {ARC-KV: Amortizing Anchor Search for Reconstruction-Based KV Cache
  Compaction}, 2026.
\newblock URL \url{https://arxiv.org/abs/2609.36835}.

\bibitem[Tian et~al.(2025)Tian, Wang, Peng, Yuan, Wang, Yi, Liu, Cui, and
  Yang]{tian2025keepkv}
Yuxuan Tian, Zihan Wang, Yebo Peng, Aomufei Yuan, Zhiming Wang, Bairen Yi, Xin
  Liu, Yong Cui, and Tong Yang.
\newblock {KeepKV: Achieving Periodic Lossless KV Cache Compression for
  Efficient LLM Inference}, 2025.
\newblock URL \url{https://arxiv.org/abs/2504.09936}.

\bibitem[Wagner et~al.(2026)Wagner, Roux, Zimmer, and
  Pokutta]{wagner2026kvsurgeon}
Moritz Wagner, Christophe Roux, Max Zimmer, and Sebastian Pokutta.
\newblock {The KV-Surgeon: Compensation-Aware Offline KV-Cache Compaction},
  2026.
\newblock URL \url{https://arxiv.org/abs/2610.25933}.

\bibitem[Wan et~al.(2024)Wan, Wu, Zhang, Xin, Tao, Zhu, Wang, Luo, Xiong, Wang,
  and Zhang]{wan2024d2o}
Zhongwei Wan, Xinjian Wu, Yu~Zhang, Yi~Xin, Chaofan Tao, Zhihong Zhu, Xin Wang,
  Siqi Luo, Jing Xiong, Longyue Wang, and Mi~Zhang.
\newblock {D2O: Dynamic Discriminative Operations for Efficient Long-Context
  Inference of Large Language Models}, 2024.
\newblock URL \url{https://arxiv.org/abs/2406.13035}.

\bibitem[Wang et~al.(2024)Wang, Jin, Yu, and Zhang]{wang2024kvmerger}
Zheng Wang, Boxiao Jin, Zhongzhi Yu, and Minjia Zhang.
\newblock {Model Tells You Where to Merge: Adaptive KV Cache Merging for LLMs
  on Long-Context Tasks}, 2024.
\newblock URL \url{https://arxiv.org/abs/2407.08454}.

\bibitem[Wu et~al.(2025)Wu, Guo, Ma, Sun, Wang, Matusik, and
  Long]{wu2025flashbias}
Haixu Wu, Minghao Guo, Yuezhou Ma, Yuanxu Sun, Jianmin Wang, Wojciech Matusik,
  and Mingsheng Long.
\newblock {FlashBias: Fast Computation of Attention with Bias}, 2025.
\newblock URL \url{https://arxiv.org/abs/2505.12044}.

\bibitem[Wu et~al.(2026)Wu, Lin, Chen, Lu, Zeng, Zhan, Zheng, and
  Kim]{wu2026semanticache}
Shunlong Wu, Hai Lin, Shaoshen Chen, Tingwei Lu, Yongqin Zeng, Shaoxiong Zhan,
  Hai-Tao Zheng, and Hong-Gee Kim.
\newblock {SemantiCache: Efficient KV Cache Compression via Semantic Chunking
  and Clustered Merging}, 2026.
\newblock URL \url{https://arxiv.org/abs/2603.14303}.

\bibitem[Xiao et~al.(2024)Xiao, Tian, Chen, Han, and
  Lewis]{xiao2024streamingllm}
Guangxuan Xiao, Yuandong Tian, Beidi Chen, Song Han, and Mike Lewis.
\newblock {Efficient Streaming Language Models with Attention Sinks}, 2024.
\newblock URL \url{https://arxiv.org/abs/2309.17453}.

\bibitem[Yang and Li(2026)]{yang2026certkv}
Chiwun Yang and Xiaoyu Li.
\newblock {Beyond Sparse Weights: When Is Attention Compressible?}, 2026.
\newblock URL \url{https://arxiv.org/abs/2608.21541}.

\bibitem[Ye et~al.(2025)Ye, Chen, Lai, Lin, Zhang, Wang, Chen, Kasikci, Grover,
  Krishnamurthy, and Ceze]{ye2025flashinfer}
Zihao Ye, Lequn Chen, Ruihang Lai, Wuwei Lin, Yineng Zhang, Stephanie Wang,
  Tianqi Chen, Baris Kasikci, Vinod Grover, Arvind Krishnamurthy, and Luis
  Ceze.
\newblock {FlashInfer: Efficient and Customizable Attention Engine for LLM
  Inference Serving}, 2025.
\newblock URL \url{https://arxiv.org/abs/2501.01005}.

\bibitem[Zhan et~al.(2026)Zhan, Chen, and Shang]{zhan2026reskv}
Yuhang Zhan, Lisi Chen, and Shuo Shang.
\newblock {ResKV: Reconstructing Omitted Attention Contributions for
  Fixed-Budget KV Cache Compression}, 2026.
\newblock URL \url{https://arxiv.org/abs/2607.29591}.

\bibitem[Zhang et~al.(2023)Zhang, Sheng, Zhou, Chen, Zheng, Cai, Song, Tian,
  R{\'e}, Barrett, Wang, and Chen]{zhang2023h2o}
Zhenyu Zhang, Ying Sheng, Tianyi Zhou, Tianlong Chen, Lianmin Zheng, Ruisi Cai,
  Zhao Song, Yuandong Tian, Christopher R{\'e}, Clark Barrett, Zhangyang Wang,
  and Beidi Chen.
\newblock {H$_2$O: Heavy-Hitter Oracle for Efficient Generative Inference of
  Large Language Models}, 2023.
\newblock URL \url{https://arxiv.org/abs/2306.14048}.

\bibitem[Zweiger et~al.(2026)Zweiger, Fu, Guo, and Kim]{zweiger2026am}
Adam Zweiger, Xinghong Fu, Han Guo, and Yoon Kim.
\newblock {Fast KV Compaction via Attention Matching}, 2026.
\newblock URL \url{https://arxiv.org/abs/2602.16284}.

\end{thebibliography}
